\documentclass[10pt,twocolumn,letterpaper]{article}

\usepackage[pagenumbers]{iccv} 
\usepackage{microtype}
\usepackage{graphicx}
\usepackage{booktabs}
\usepackage{amssymb}
\usepackage{algorithm,algorithmic}
\usepackage{amsmath}
\usepackage{tabularx}
\usepackage{multirow}
\usepackage{color}
\usepackage{comment}
\usepackage{listings}
\usepackage{multicol}
\usepackage{subcaption} 
\definecolor{iccvblue}{rgb}{0.21,0.49,0.74}
\usepackage[pagebackref,breaklinks,colorlinks,allcolors=iccvblue]{hyperref}

\title{Understanding Attention Mechanism in Video Diffusion Models}

\author{Bingyan Liu$^{1,2}$, Chengyu Wang$^2$\thanks{Correspondence to:chengyu.wcy@alibaba-inc.com}, Tongtong Su$^{3,2}$,Huan Ten$^{1,2}$, Jun Huang$^2$,Kailing Guo$^1$,Kui Jia$^4$\\
$^1$South China University of Technology,
$^2$ Alibaba Group,
$^3$ Zhejiang University,\\
$^4$The Chinese University of Hong Kong, Shenzhen\\
}

\begin{document}
\maketitle

\begin{abstract}
Text-to-video (T2V) synthesis models, such as OpenAI's Sora, have garnered significant attention due to their ability to generate high-quality videos from a text prompt. In diffusion-based T2V models, the attention mechanism is a critical component. However, it remains unclear what intermediate features are learned and how attention blocks in T2V models affect various aspects of video synthesis, such as image quality and temporal consistency. In this paper, we conduct an in-depth perturbation analysis of the spatial and temporal attention blocks of T2V models using an information-theoretic approach.
Our results indicate that temporal and spatial attention maps affect not only the timing and layout of the videos but also the complexity of spatiotemporal elements and the aesthetic quality of the synthesized videos. Notably, high-entropy attention maps are often key elements linked to superior video quality, whereas low-entropy attention maps are associated with the video's intra-frame structure.
Based on our findings, we propose two novel methods to enhance video quality and enable text-guided video editing. These methods rely entirely on lightweight manipulation of the attention matrices in T2V models. The efficacy and effectiveness of our methods are further validated through experimental evaluation across multiple datasets.
\end{abstract}
\vspace{-.5em}    
\section{Introduction}
\label{sec:intro}
Recently, large-scale text-to-video (T2V) synthesis models, such as Gen-3~\cite{Gen3}, HuanyuanVideo ~\cite{li2024hunyuan}, and Sora~\cite{sora}, have demonstrated outstanding performance in video synthesis. To our knowledge, T2V models in the open-source community predominantly utilize two main structural approaches. The first, exemplified by AnimateDiff~\cite{guo2023animatediff,lin2024animatediff}, VideoCrafter2~\cite{videocrafter2}, and LaVie~\cite{lavie}, employs pre-trained text-to-image (T2I) models~\cite{LDM} to initialize T2V model checkpoints, subsequently adding specialized motion modules to concentrate on learning the temporal dynamics of videos. In contrast, alternative strategies move away from pre-trained T2I models by redesigning the architecture and training the model from scratch. A notable example, CogVideoX~\cite{Cogvideo5B}, introduces a 3D causal VAE and expert transformer blocks equipped with 3D full attention.

However, current T2V models face several challenges in practical applications. Firstly, achieving high imaging quality for T2V models remains a significant challenge compared to state-of-the-art T2I models~\cite{LDM,chen2023pixart}. Secondly, there is insufficient research exploring the potential of T2V models in video editing~\cite{tokenflow,wu2023tune,slicedit}. These models often lack the capability to semantically modify a given video at either a local or global scale using prompts. In practice, even minor adjustments to the prompt can result in a video that diverges substantially from the original intent. More importantly, compared to attention blocks in diffusion-based T2I models~\cite{Liu_2024_CVPR,PIE,hertz2022prompt}, research on the role of attention mechanisms within video diffusion models (VDMs), and their influence on video generation outcomes, remains limited.

In this paper, our objective is to gain comprehensive insight into attention mechanisms of diffusion-based T2V models. Specifically, we propose perturbing attention maps using an identity matrix \(I\) and a uniform matrix \(U\) (i.e., a uniform row-stochastic matrix where each row sums to 1) and pose the fundamental question: \emph{How does attention in T2V models affect the outcomes of video synthesis?} To address this question, we propose an explainable exploration method that employs maximum and minimum perturbations on spatial and temporal attention to investigate the impact of attention maps on resulting videos. Our findings are as follows: 
1. Attention maps are crucial for elucidating the functioning of VDMs. Attention maps similar to a uniform matrix (with high entropy) are essential for determining the quality of video frames, whereas low-entropy attention maps encompass key structural layout information.
2. Spatial attention layers are vital for successful video editing, as they preserve the original video's layout information and shape details at the frame level.
3. Temporal attention layers play a pivotal role in effective video synthesis. Attention in temporal layers dictates the dynamic properties of the video and influences the structural features within each video frame.

Based on our findings, we introduce \emph{information entropy-driven adaptation} (IE-Adapt), an efficient and training-free technique for manipulating the attention maps of diffusion-based T2V models. By performing lightweight operations on selected attention maps based on their entropy, we effectively execute the following tasks with two novel methods: (1) enhancing video quality during the diffusion denoising process and (2) facilitating text-guided video editing purely based on the T2V backbone. Our approach not only optimizes or modifies the generation outcomes but also offers a new perspective for further exploring the role of attention mechanisms in T2V models.

The contributions of our paper are as follows:
\begin{itemize}
    \item We introduce a novel perturbation approach that facilitates both exploration and interpretative analysis of the spatial and temporal attention layers within VDMs.
    \item Our research reveals that attention maps significantly influence various facets of generated videos: low-entropy attention maps capture the structural and dynamic attributes of videos, while high-entropy attention maps can affect the quality of generated videos.
    \item We propose entropy-based adaptation for enhancing both video synthesis quality and video editing capabilities. Experimental results demonstrate the effectiveness of our method, underscoring the value of our work.
\end{itemize}

\section{Preliminaries and Related Work}

\noindent\textbf{Diffusion models.} Diffusion probabilistic models (DPMs) \cite{ho2020denoising} are a category of generative models designed to approximate data distribution through progressive denoising. By integrating additional guiding signals, such as text-based conditions \( c \), DPMs can also learn a conditional distribution and generate images guided by these signals, exemplified by Stable Diffusion~\cite{LDM}. Beginning with a Gaussian noise \( x_T \sim \mathcal{N}(0, I) \), with a noise schedule \(\bar\alpha=\prod_{1}^{t}(1-\beta_{t})\) according to a variance schedule \(\beta_{1},\dots, \beta_{t}\), the conditional diffusion model \( \epsilon_{\theta} \) incrementally removes noise until it yields a clean image \( x_0 \). The denoising process using the DDIM sampler \cite{songdenoising} is defined as:
\begin{equation}
    x_{t-1} = \sqrt{\overline\alpha_{t-1}} \hat{x}_0 + \sqrt{1-\overline\alpha_{t-1}} \epsilon_{\theta}\left(x_{t}, t, c\right),
\label{eq:x_sample}
\end{equation}
where \(\hat{x}_0\) is the predicted image latent at timestep \(t\): $\hat{x}_0 = \frac{x_{t} - \sqrt{1 - \overline{\alpha}_{t}} \epsilon_{\theta}(x_{t}, t, c)}{\sqrt{\overline{\alpha}_{t}}}$.

\noindent\textbf{Video diffusion models.} To our knowledge, there is still no consensus on the optimal design for spatial and temporal attention mechanisms in VDMs. Some VDMs share the same latent space with Stable Diffusion~\cite{guo2023animatediff,videocrafter2}. These models extend Stable Diffusion to the video domain by adding temporal self-attention to model the motion pattern between video frames. AnimateDiff~\cite{guo2023animatediff} freezes the pretrained spatial module, training only the additional temporal module. This aims to disentangle motion from spatial information. Ideally, the temporal module should solely capture motion information. However, in practice, when replacing the spatial base model with a T2I model, both image quality and motion quality degrade significantly~\cite{videocrafter2}. This indicates that there is entanglement of spatial and temporal information, which motivates us to explore how attention aggregates. Another stream of VDMs is trained from scratch~\cite{Cogvideo5B,xu2024easyanimate}, mixing large amounts of images and videos as training data. They adopt DiT (Diffusion Transformer)~\cite{peebles2023scalable} to replace convolutional blocks in U-Net, demonstrating enhanced generative capabilities when the model is scaled up. For example, CogVideoX~\cite{Cogvideo5B} uses full 3D attention in their DiT blocks.

\noindent\textbf{Attention layers.} Whether the denoiser architecture is built on U-Net or transformer, attention serves as a fundamental component, both between text and images and within images themselves. Understanding how attention aggregates and propagates features facilitates many downstream tasks such as editing~\cite{FPE,tumanyan2023plug} and enhancement~\cite{si2024freeu,ahn2024self} in a zero-shot manner. These works mainly focus on self-attention, since editing authentic images on cross-attention layers can result in editing failures~\cite{FPE,tumanyan2023plug}. Specifically, self-attention linear projections \(\ell_{q}\), \(\ell_{k}\), and \(\ell_{v}\) map the input image features \(z_{t}\) onto query, key, and value matrices, \(Q = \ell_{q}(\phi(z_{t})), \quad K = \ell_{k}(\phi(z_{t})), \quad V = \ell_{v}(\phi(z_{t}))\), respectively. The attention map \(A\) is defined as:
\(A = \text{Softmax}\left(\frac{Q  K^{\mathsf{T}}}{\sqrt{d}}\right)\)
where \(d\) is the dimension of keys and queries. The output is defined as the fused feature from the projected features according to spatial similarity, denoted as \(\hat{\phi}(z_{t}) = A  V\).

\noindent\textbf{Video processing.} Recent video processing tasks, including editing, style transfer, and quality enhancement, primarily utilize T2I models on a frame-by-frame basis~\cite{yang2023rerender,yang2024fresco,tokenflow,zhang2023controlvideo,qi2023fatezero}. These methods incorporate cross-frame attention layers instead of self-attention in U-Net to maintain temporal consistency. However, they often struggle with preserving local consistency, leading to flickering artifacts. Rerender-A-Video~\cite{yang2023rerender,yang2024fresco} leverages optical flow from source videos to warp and integrate latent features into edited outputs, whereas TokenFlow~\cite{tokenflow} uses inter-frame feature matching and reference frame feature propagation. Although these methods effectively enforce local consistency, they require source videos with simple motion patterns for accurate estimation of consistency constraint~\cite{Liang_2024_CVPR,hu2023videocontrolnet}. Advancements in VDMs, particularly their temporal attention, have opened avenues for direct video editing. Since temporal attention also encompasses spatial information, further research is needed to improve understanding of attention in VDMs.

\section{Perturbation Analysis on Attention Maps}
\label{pert_atten_map}
In this section, we analyze how spatial and temporal attention maps in VDMs contribute to video synthesis.

\subsection{Design of Perturbation Analysis}

Despite the existence of extensive research on VDMs, the effects of spatial and temporal attention in these models remain unclear. Inspired by black-box explainability techniques~\cite{interpre_clip}, we propose an approach that perturbs attention maps and uses feature quantification for analysis.

Previous research has revealed that attention maps in Stable Diffusion are closely related to the semantic and structural information of images~\cite{FPE,hertz2022prompt,tumanyan2023plug}. Based on these observations, we propose perturbing the attention map \(A\). One approach replaces \(A\) with the identity matrix \(I\), effectively bypassing attention by using only \(V\).Another approach substitutes \(A\) with a uniform matrix \(U\), thereby removing the original influence of the attention map. Here, \(A\), \(U\), and \(I\) are all \(N \times N\) matrices. 
We utilize entropy \(\mathcal{H}\) and energy \(\mathcal{E}\) to demonstrate their validity, defined as:
$\mathcal{H}(A) = -\sum_{i} A_{i} \log A_{i}$ and 
$\mathcal{E}(A  V) = \sum_{i} \left( (A  V)_{i} \right)^2$
where \(\mathcal{H}(A)\) denotes the information entropy of \(A\), with \(A_i\) as the attention weight for element \(i\), measuring distribution uncertainty.
According to information theory \cite{LESNE_2014,gray2011entropy}, the identity matrix \(I\) has the lowest entropy, while \(U\) has the highest. From an energy perspective yields similar results, expressed as follows:
\begin{equation}
\begin{aligned}
0 = \mathcal{H}(I) \leq \mathcal{H}(A) \leq \mathcal{H}(U) \\
\mathcal{E}(U) \leq \mathcal{E}(A) \leq \mathcal{E}(I) = N
\end{aligned}
\end{equation}
Empirically, the attention map in deep models smooths the values through weighting, resulting in a reduction in the energy of the computed weighted average (\(A  V\)). 
We assume that the upper bound of \(\mathcal{E}(AV)\) is close to that of \(\mathcal{E}(V)\), while the lower bound is represented by \(\mathcal{E}(UV)\). 
The energy \(\mathcal{E}(A  V)\) indicates the magnitude of features after applying \(A\) to \(V\). \( \mathcal{E}(AV) \) exhibits a high probability of falling within the interval \([\mathcal{E}(UV), \mathcal{E}(IV)]\). As shown in Figure~\ref{fig:entropy_and_energy}, across various attention layers,  \( \mathcal{E}(AV) \) is entirely bounded by \( \mathcal{E}(UV) \) and \( \mathcal{E}(IV) \).
Thus, the aforementioned method effectively replaces the attention map \(A\) with its entropy and energy-level upper and lower bounds. The perturbed attentions are represented as:
\({\hat{\phi}_{I}}(z^{l}_{t}) = I  V\) and \({\hat{\phi}_{U}}(z^{l}_{t}) = U  V\).

To study the effect of different attention maps, we employ maximum and minimum perturbation techniques. In the experiments, all tests involve replacing attention maps with identity and uniform matrices over the video synthesis benchmark VBench~\cite{VBench}, containing 93 video generation prompts. For each prompt, \(2n\) videos will be generated where \(n\) is the number of attention layers of the T2V model.
We measure the differences in structural similarity, aesthetics, and temporal consistency of output videos before and after perturbation. Components with larger impacts should lead to more significant fluctuations in evaluation metrics. Our primary analysis focuses on the U-Net of AnimateDiff \cite{guo2023animatediff} and also studies other VDMs such as CogVideoX~\cite{Cogvideo5B,cogvideo}.

\begin{figure}[ht]
\vspace{-.5em}
\centering
\includegraphics[width=0.48\textwidth]{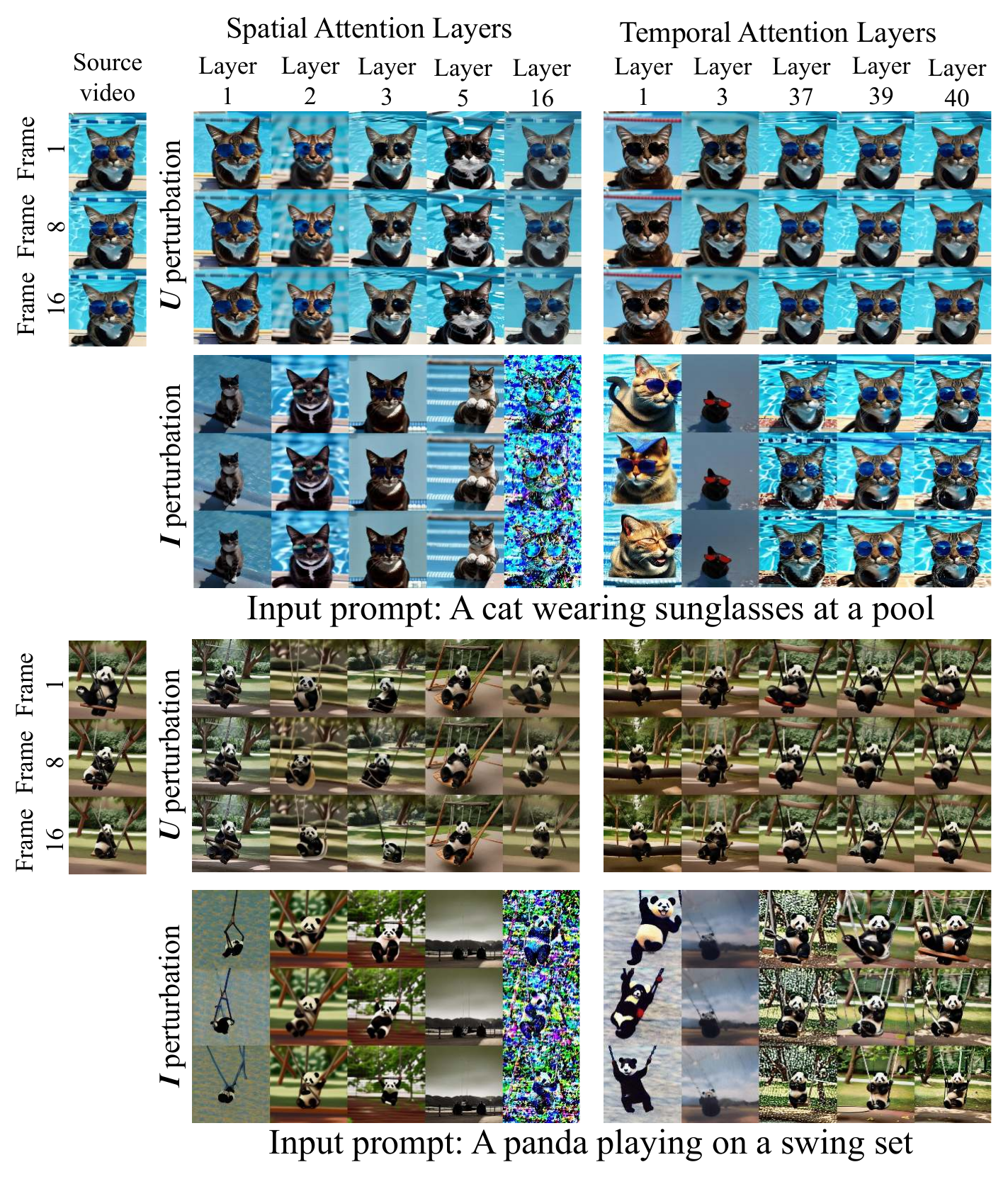}
\vspace{-1.25em}
\caption{Perturbation results on AnimateDiff. $I$ perturbation refers to replacing $A$ with $I$ at the $L$-th layer, while $U$ perturbation substitutes it with a uniform matrix. The input prompts are selected from VBench. Perturbation is conducted on both spatial and temporal attention layers, where only one layer is perturbed at a time.
}
\label{fig:perturbed_video_case}
\vspace{-.5em}
\end{figure}

\begin{figure}[ht]
\centering
\vspace{-.5em}
\includegraphics[width=0.48\textwidth]{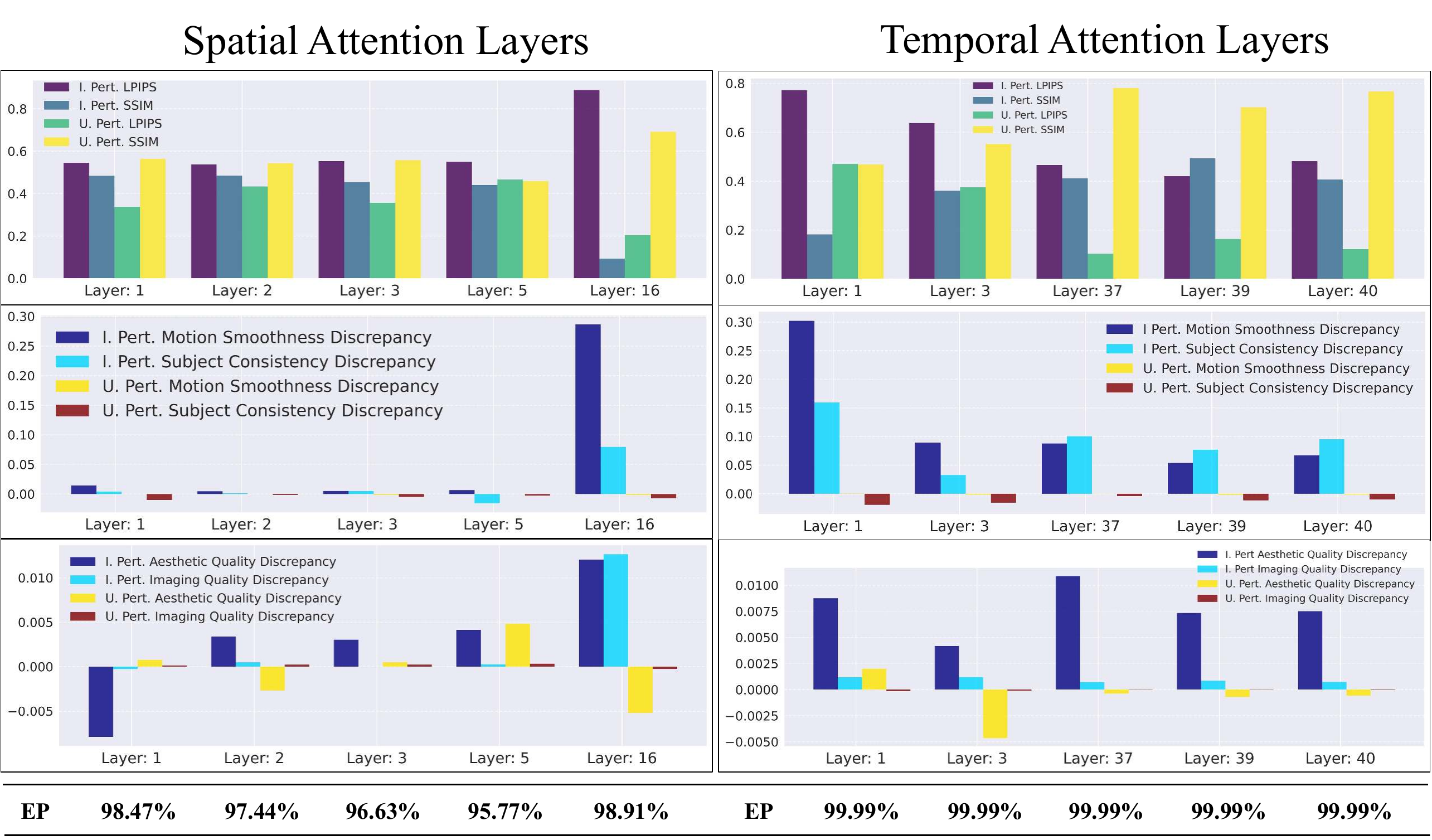}
\caption{Histogram of attention map perturbation results on AnimateDiff. From top to bottom: structural measure, temporal consistency difference before and after perturbation, and shift in aesthetics. Smaller values for LPIPS indicate better performance, while larger values are preferred for the other metrics. EP: entropy percentage.}
\label{fig:Perturbed_histogram}
\vspace{-1.5em}
\end{figure}

\subsection{Perturbation on Spatial Attention Layers}

\noindent\textbf{Spatial attention maps control the structural and imaging information of the generated videos.} When these attention maps are perturbed by our approach, the intra-frame structure, aesthetics, and inter-frame consistency of the video are altered. Figures~\ref{fig:perturbed_video_case} and~\ref{fig:Perturbed_histogram} present the perturbation results. Our results show that spatial attention layers at both the beginning and end of the U-Net network have the largest impact on the video, with the final layer being the most affected.
Specifically, spatial attention maps from different layers capture different types of information. At the initial and final attention layers of U-Net, replacing the layer with \(I\) significantly affects the intra-frame image structure of the video. It is further observed in Figure~\ref{fig:Perturbed_histogram} that using \(I\) increases the Learned Perceptual Image Patch Similarity (LPIPS) index~\cite{LPIPS} and decreases the Structural Similarity (SSIM) index~\cite{SSIM}. Conversely, replacing these layers with \(U\) does not significantly impact the image structure. However, replacing attention maps with \(I\) near the bottleneck of U-Net does not have a significant effect, while replacing them with \(U\) produces the opposite result.

As for inter-frame consistency, our perturbation results show that replacing spatial attention maps with \(I\) or \(U\) generally has little impact, except for the final 16th U-Net layer. In this layer, replacing the spatial attention maps with \(I\) causes a sharp decline in inter-frame consistency. \textbf{Therefore, in contrast to common belief, the spatial attention maps of the final layer also affect temporal consistency.}

\subsection{Perturbation on Temporal Attention Layers}

\noindent\textbf{Most noticeably, temporal attention maps reflect the amplitude of motion in video synthesis.} As shown in Figures~\ref{fig:perturbed_video_case} and \ref{fig:Perturbed_histogram}, perturbing temporal attention maps affects the video's motion patterns and temporal consistency. When the attention map is highly similar to \(U\), the generated video tends to be static; conversely, when it is closer to \(I\), although the amplitude of motion increases, the inter-frame consistency may decrease significantly. Regarding the temporal consistency of the video, altering the initial and final attention layers of the U-Net with \(I\) has a more pronounced impact.

\noindent\textbf{A somewhat surprising finding is that temporal attention maps also affect the video's aesthetics and imaging quality.} This impact is considerable, especially in the initial and final attention layers of U-Net. Our results show that replacing temporal attention maps with \(I\) leads to a decline in the quality and aesthetics of the video images. \textbf{Although the temporal layer carries a minor effect on the layout structure of video images, unreasonable perturbations may yet result in significant changes in the structure.} Similar to spatial attention blocks, inserting \(I\) at the initial and final attention layers of U-Net significantly impacts the intra-frame image structure of the video. Replacing the matrix with \(I\) in attention maps near the bottleneck of U-Net does not have a significant effect, whereas substituting it with \(U\) produces the opposite result.

\subsection{Information Theoretic View of Attention Maps}
It is evident that both spatial and temporal attention are entangled in video synthesis. We further claim that the \emph{information theoretic view} of attention, regardless of being spatial or temporal, is the key to understanding attention.

\begin{figure}[ht]
\centering
\vspace{-.5em}
\includegraphics[width=0.475\textwidth]{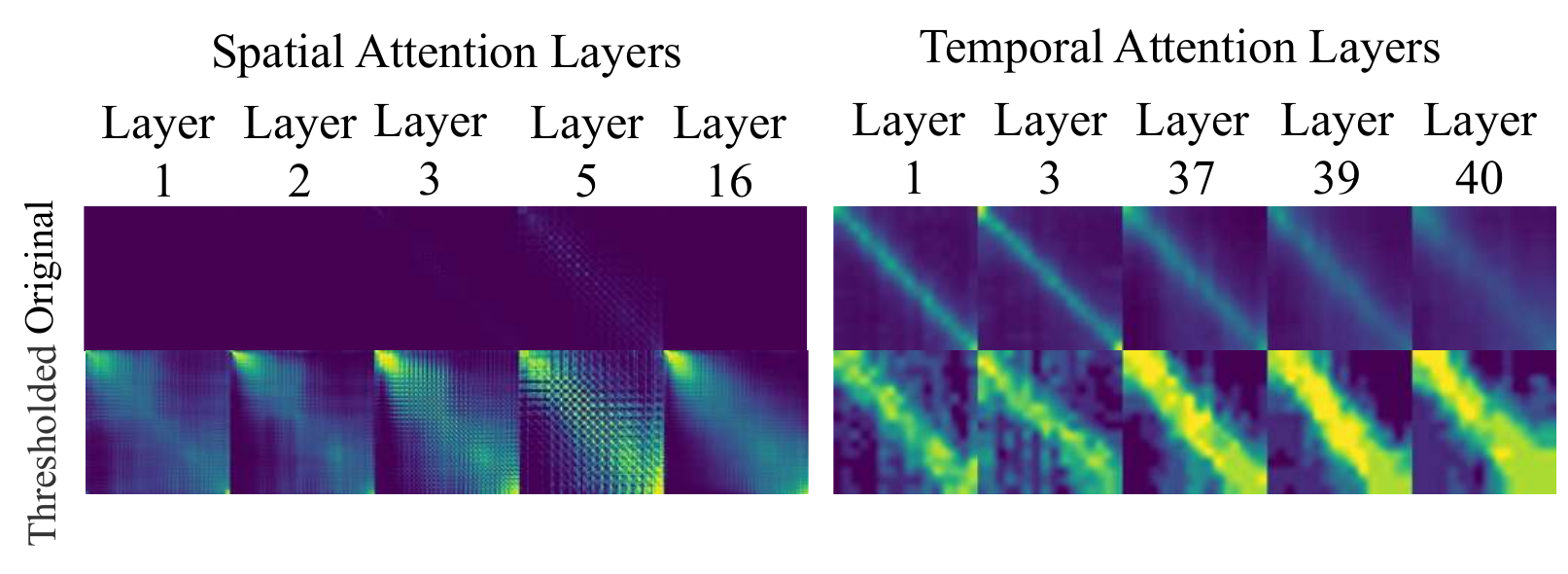}
\vspace{-1em}
\caption{Visualization results of attention maps in the spatial and temporal layers of AnimateDiff. ``Original'' refers to the visualization of the original attention map. ``Threshold'' indicates the visualization after using a threshold, allowing for a clearer view of the distribution of attention map values.}
\label{fig:attenmap}
\vspace{-.5em}
\end{figure}

\begin{figure}[ht]
\centering
\vspace{-.5em}
\includegraphics[width=0.49\textwidth]{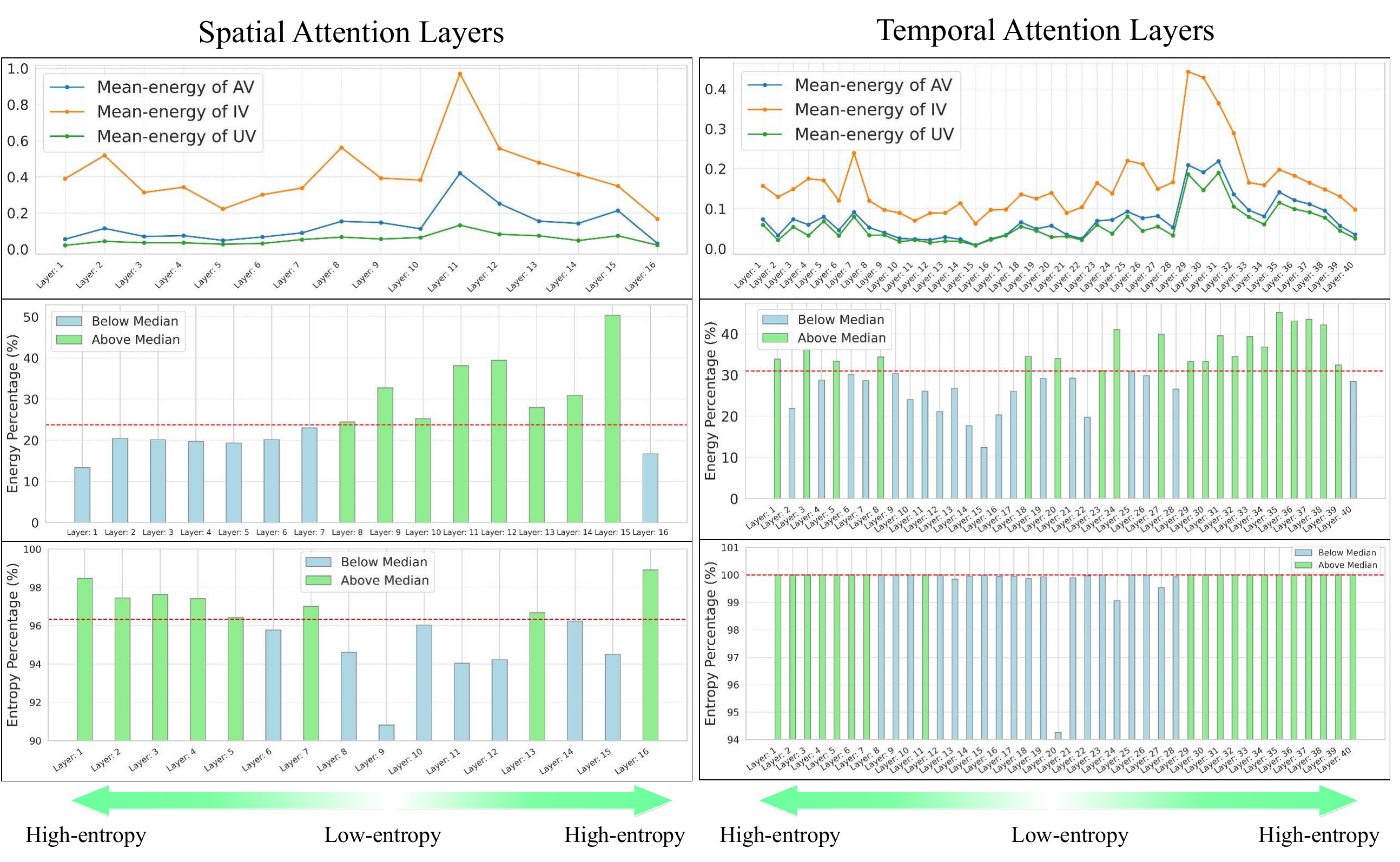}
\caption{Entropy and energy of attention layers in AnimateDiff. Top to bottom: energy values, energy proportion, information entropy proportion.
The red line divides proportion values into top 50$\%$ and bottom 50$\%$. Information entropy is normalized due to varying upper bounds of different layer sizes.}
\label{fig:entropy_and_energy}
\vspace{-1em}
\end{figure}

As shown in Figure~\ref{fig:entropy_and_energy}, in spatial attention layers, the energy values of nearly all attention outputs are close to the lower threshold of the layer's energy. Specifically, these attention maps tend to approximate the uniform matrix \(U\). These attention maps, visualized without thresholding, appear as dark images, as shown in Figure~\ref{fig:attenmap}. From an information-theoretic perspective, the entropy of attention maps shows an inverse association with the energy of the output results derived from the maps. Consequently, if these attention maps are replaced with \(I\), there would be an increase in the output energy. Such a modification could potentially disrupt the original image's structural features, leading to a change in image structure, as evidenced by low SSIM and high LPIPS post-perturbation.
Consider the U-Net architecture of the diffusion model. Attention maps of the starting and final layers are closer to \(U\) than those of the bottleneck layer. Thus, replacing bottleneck layer attention with \(I\) leads to an increase in energy and changes the image's structural layout. This observation suggests that attention maps in middle layers contain more structural information.

As for temporal attention layers, our statistics reveal that the entropy percentage of more than 70\% of attention maps in AnimateDiff is close to 100\%. Intuitively, a temporal attention map close to \(U\) indicates that the video frame associated with the attention is highly static. In contrast, the closer the attention map is to the identity matrix \(I\), the richer the generated video frames usually are in motion. In other words, the relationship between frames will be better captured. Based on this observation, \textbf{high entropy in attention maps often leads to key elements that are associated with better imaging quality and inter-frame consistency.}

\begin{figure}[ht]
\centering
\includegraphics[width=0.45\textwidth]{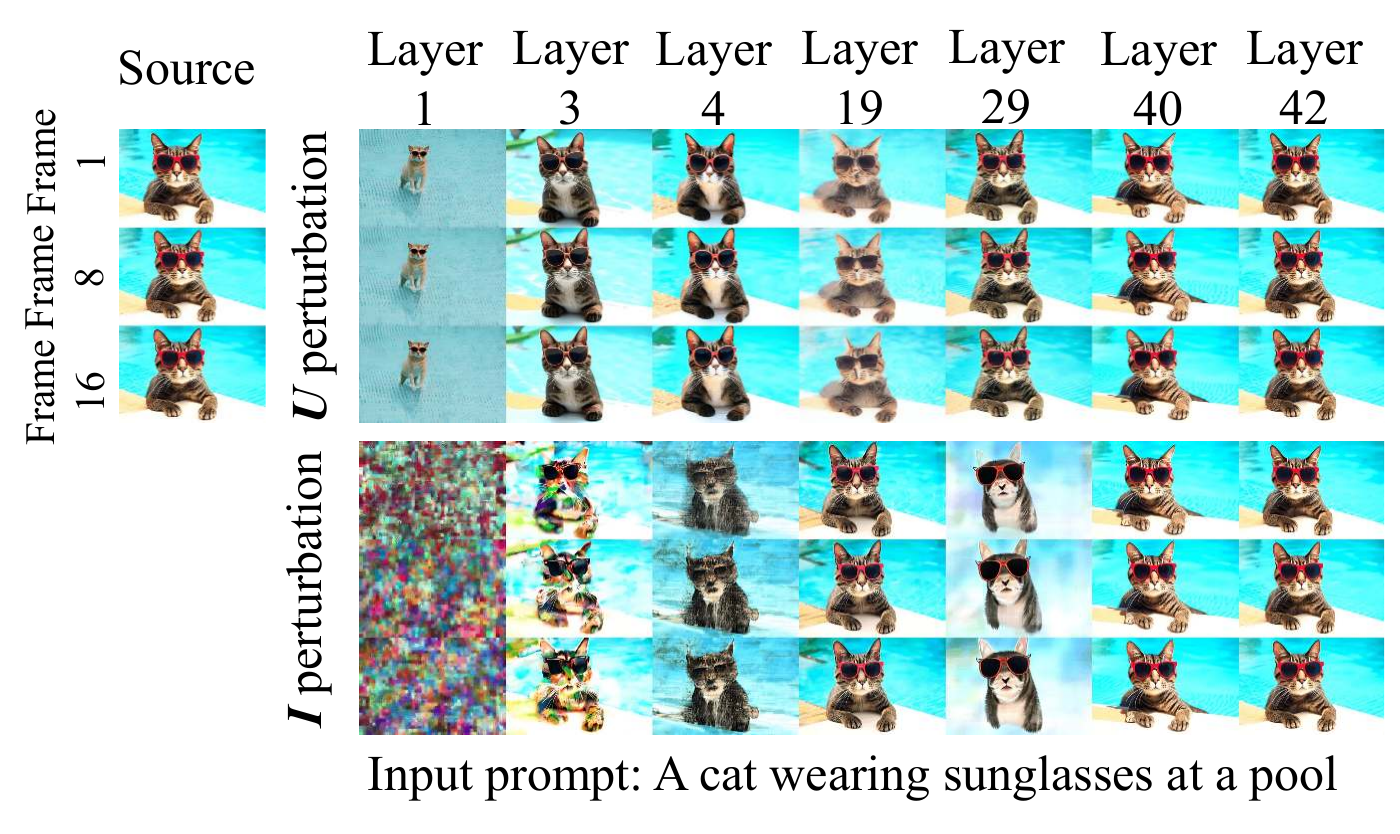}
\caption{Perturbation results in CogVideoX-5B.}
\vspace{-1.5em}
\label{fig:cogvideo}
\end{figure}

\begin{figure}[ht]
\centering
\includegraphics[width=0.475\textwidth]{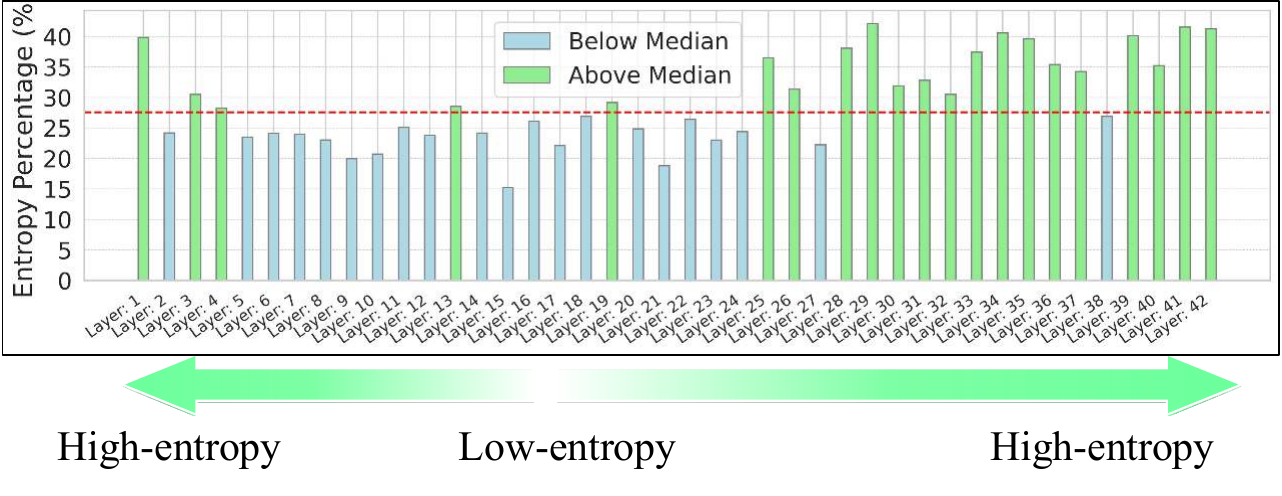}
\caption{Attention map information entropy in CogVideoX-5B.}
\label{fig:cogvideo_entropy}
\vspace{-1em}
\end{figure}

\subsection{Observations from Other Models}

We conduct further perturbation on a different structure of VDM, CogVideoX~\cite{Cogvideo5B}. As shown in Figure~\ref{fig:cogvideo} and Figure~\ref{fig:cogvideo_entropy}, similar to the U-Net architecture, attention maps of CogVideoX exhibit higher information entropy in the initial and final layers of the network compared to the middle layers. Nevertheless, there are a few differences: CogVideoX is more sensitive to perturbations in the shallow parts but exhibits greater robustness to perturbations in the final layer. We refer readers to the supplementary materials for details.

\section{Applications}

Based on our analysis, we propose two algorithms to improve the effectiveness of video synthesis and video editing.

\begin{figure}[ht]
\centering
\vspace{-.5em}
\includegraphics[width=0.4\textwidth]{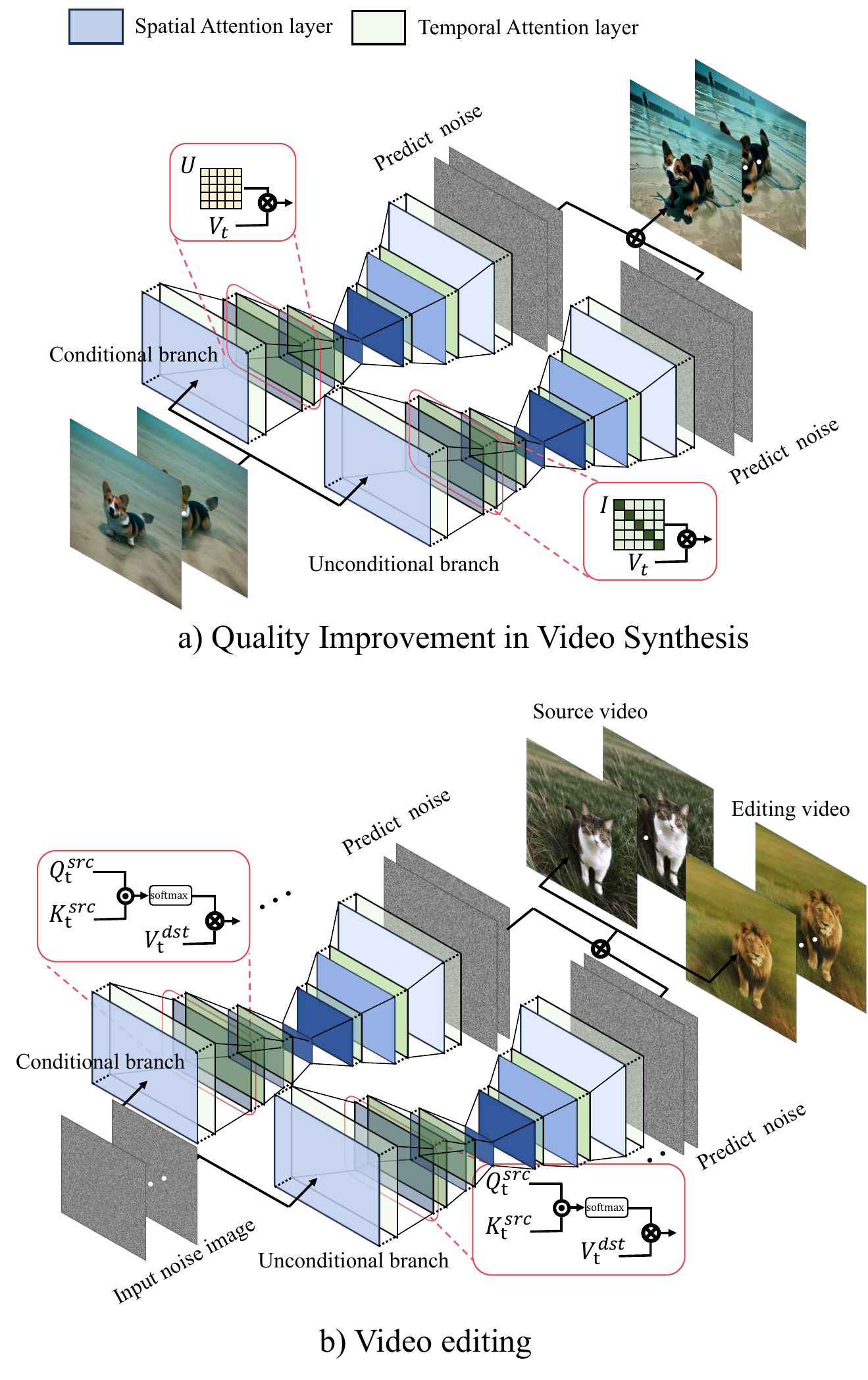}
\vspace{-1em}
\caption{Two applications based on IE-Adapt.}
\vspace{-1em}
\label{fig:method}
\end{figure}

\subsection{Quality Improvement in Video Synthesis}

When we aim to improve video quality during the denoising process, attention maps with higher entropy should be prioritized by replacing them with \(U\) in the positive branch and with \(I\) in the negative branch. We refer to this as \emph{information entropy targeted adaptation} (IE-Adapt). The process for improving video quality is as follows: for \(n\) layers of attention maps \(A_{1:n}\) in model \(\mathcal{F}\), we identify the layer \(i\) with maximum entropy and record it in our registry:
\begin{equation}
    \mathcal{F}^* \leftarrow \text{Register}(\mathcal{F}, i), i = \mathop{\arg\max}\limits_{i \in [1,n]} \mathcal{H}(A_i).
\end{equation}
In \(\mathcal{F}^*\), when the forward pass reaches the registered layer, original attention maps \(A_{i}\) are replaced with \(I\) or \(U\), depending on the branch. As illustrated in Figure~\ref{fig:method}, these two additional branches yield the outputs \(\epsilon_{\theta}(x,c,I)\) and \(\epsilon_{\theta}(x,c,U)\). Along with the original branch using attention maps \(A_{1:n}\), we obtain the following four estimated noises:
\begin{equation}
\begin{aligned}
    \epsilon_{\theta}(x,c,A), \epsilon_{\theta}(x,\phi,A) &\leftarrow \mathcal{F}(z,t,c), \\
    \epsilon_{\theta}(x,c,U), \epsilon_{\theta}(x,\phi,I) &\leftarrow \mathcal{F}^*(z,t,c).
\end{aligned}
\end{equation}

Analogous to the Classifier-Free Guidance (CFG) technique~\cite{ho2022classifier}, which applies a CFG weight \(\omega\) to accentuate the text condition \(c\) over the unconditional branch with the null text \(\phi\), \((\epsilon_{\theta}(x,c,A) - \epsilon_{\theta}(x,\phi,A))\), we introduce additional guidance leveraging imaging quality enhancement or degradation based on the aforementioned analysis: \(\epsilon_{\theta}(x,c,U) \succ \epsilon_{\theta}(x,c,A) \succ \epsilon_{\theta}(x,c,I)\). To disentangle this from the original text condition, we separately add a term:
\begin{equation}
\begin{split}
\epsilon = & \ \epsilon_{\theta}(x,\phi,A) + \omega \cdot (\epsilon_{\theta}(x,c,A) - \epsilon_{\theta}(x,\phi,A))\\
& + \lambda \cdot (\epsilon_{\theta}(x,c,+) - \epsilon_{\theta}(x,c,-))
\end{split}
\label{eq:CFG_Ours}
\end{equation}
where \(\lambda\) is independent of \(\omega\) to flexibly control aesthetic enhancements beyond the text condition. Positive and negative terms, i.e., \(\epsilon_{\theta}(x,c,+)\) and \(\epsilon_{\theta}(x,c,-)\), can be chosen according to the estimation quality ranking above. We perform ablation studies across different combinations in experiments. The method is detailed in Figure~\ref{fig:method}.

\subsection{Video Editing}

For video editing, let \(V_{src}\) be the video to be edited. Our goal is to synthesize a video \(V_{dst}\) based on the target prompt \(P_{dst}\) while preserving the content and structure of the original video \(V_{src}\). Our idea is to leverage the entropy of attention maps to guide the editing process. Attention maps with higher entropy contain less structural information; modifying them could significantly alter structural and inter-frame relationships, so these maps should be preserved. Conversely, attention maps with lower entropy contain more structural information and should be utilized in the denoising process. The IE-Adapt process for video editing is expressed as: \(\mathcal{F}^* \leftarrow \text{Register}(\mathcal{F}, \mathcal{I})\) where \(\mathcal{I}\) is the index collection in which the entropy of each \(A_i\) (\(i \in \mathcal{I}\)) is among the bottom 50\% of all attention maps in the model. The choice of the entropy ratio of 50\% was based on empirical results, discussed in the supplementary material.

Algorithm 1 (see supplementary materials) presents the pseudocode of our algorithm. Unlike existing video editing methods that pre-set which attention layers to replace~\cite{tokenflow,ku2024anyv2v,Liu_2024_CVPR,qi2023fatezero}, our approach initially calculates the entropy for each layer's attention maps. This information guides the decision of whether to replace the attention maps of each layer. For videos generated by T2V models, we replace the target video's attention maps in the entropy-guided layers with those from the source video during the diffusion denoising process. For real-world video editing, we first reconstruct the real video to obtain the source video's attention maps using the inversion operation~\cite{songdenoising} . Then, during editing, we replace the targeted attention maps of the real video within the generation process of the target video.

\section{Evaluation on Two Applications}
\subsection{Experimental Settings}

For video quality improvement, we leverage the datasets in VBench~\cite{VBench}, which contain 1746 text prompts, and select 30 videos generated from the VBench prompt sets that exhibit problems such as artifacts and lack of details. For video editing, considering the scarcity of public datasets for verifying the effectiveness of text-video editing, we collect text prompts from publicly available image editing datasets created in prior research~\cite{PIE}, including the PIE-Benchmark~\cite{PIE}, with a total of 170 source-target prompts. For real video editing, we collect videos from the DAVIS dataset~\cite{Davis} and the LOVEU-TGVE dataset~\cite{TGVE}, comprising 20 text-video pairs.
For a fair comparison, we fix the number of frames in all edited videos by removing frames from the outputs of other methods. We employ the Aesthetic Score (AS)~\cite{Laion_5b} for image quality evaluation, as well as Clip Score (CS), Clip Directional Similarity (CDS)~\cite{clip,StyleGAN-NADA}, Motion Smoothness (MS), and Subject Consistency (SC)~\cite{VBench} to quantitatively analyze the experimental results of video editing using currently popular video editing algorithms. The underlying model utilized is AnimateDiff.

\subsection{Results of Video Quality Enhancement}
\begin{figure}[ht]
\centering
\includegraphics[width=0.495\textwidth]{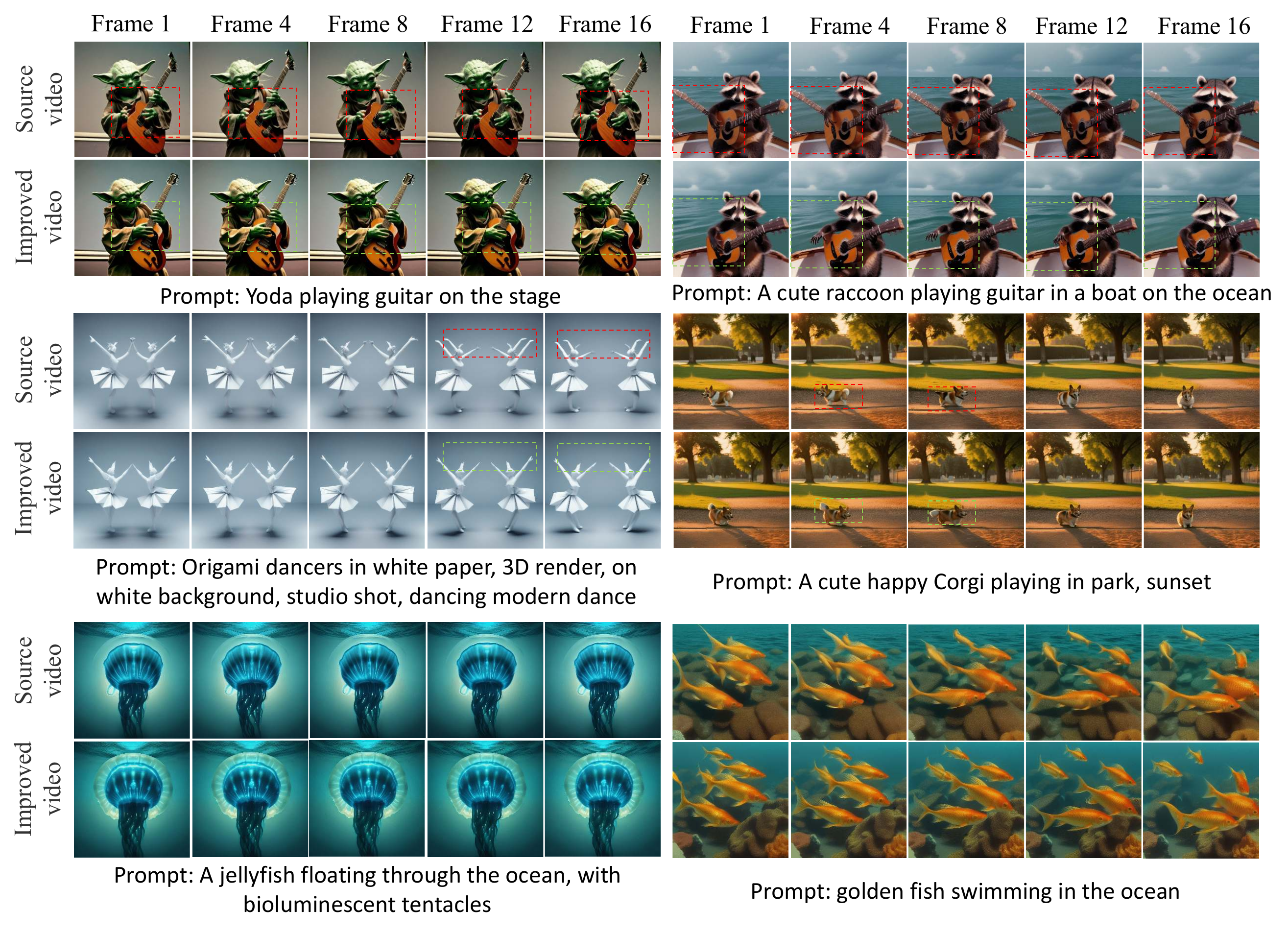}
\vspace{-1.25em}
\caption{Qualitative analysis of video quality improvement.}
\label{fig:improved_video}
\vspace{-1em}
\end{figure}

In Figure~\ref{fig:improved_video}, we present additional results highlighting the improvements in video synthesis quality. As shown in the figure, our method effectively rectifies deformities that occur during the video generation process, such as Yoda's hand, the raccoon's guitar, the dancer's hand, and the corgi's head. Additionally, our approach significantly enhances detailed information between frames, exemplified by the complex textures of the jellyfish and the goldfish. These improvements underscore the effectiveness of our method in enhancing video synthesis quality.
We then assess the video generation quality using the entire VBench dataset, which contains 1746 text prompts. Results are shown in Table~\ref{video_quality_more}. Our method improves video quality, aesthetics, and subject consistency.

\begin{table}[ht]
\centering
\setlength{\aboverulesep}{0pt}
\setlength{\belowrulesep}{0pt}
\begin{footnotesize}
\begin{tabular}{lcccc} 
\toprule
Method  & AS  & IQ & MS & SC \\  
\midrule
Original & 0.3238 & 0.1968 & \textbf{0.9777} & 0.9819 \\
Ours & \textbf{0.3295} & \textbf{0.1972} & 0.9740 & \textbf{0.9937}  \\
\bottomrule
\end{tabular}
\end{footnotesize}
\caption{Quantitative results of video quality improvement.}
\label{video_quality_more}
\vspace{-.5em}
\end{table}

We further discuss three combinations in Eq.~\ref{eq:CFG_Ours}. Specifically, the combinations can be denoted as \([A-I]\), \([U-A]\), and \([U-I]\), representing the positive term minus the negative term as the guidance for enhancement. Previous work~\cite{ahn2024self} perturbs with an identity matrix for the T2I model as the negative term, which can be aligned with \([A-I]\) in our work. Their replacement layers are selected empirically for each case. Applying this to the video diffusion model is more challenging due to its large number of temporal attention layers. As shown in Table~\ref{tab:improved_video}, all types of guidance can enhance the image quality of the video, and \([A-I]\) achieves the highest average score. Besides the average, we also focus on the potential to attain the highest level of performance. \([U-I]\) has more potential to achieve the best performance, specifically in attaining the best average rank~(\(\downarrow\)).

\begin{table}[ht]
\centering
\aboverulesep=0pt
\belowrulesep=0pt
\begin{footnotesize}
\begin{tabular}{c|cccc}
\toprule
  & Original & $[A-I]$ & $[U-A]$ & $[U-I]$ \\
\hline
AS~(Avg) $\uparrow$  & 0.6221 & \textbf{0.7023} & 0.6753 & \underline{0.6967} \\
AS~(Rank) $\downarrow$ & - & \underline{1.86} & 2.46 & \textbf{1.66}\\
\hline
\bottomrule
\end{tabular}
\end{footnotesize}
\caption{Analysis on three guidance strategies.}
\vspace{-1em}
\label{tab:improved_video}
\end{table}

\begin{figure}[ht]
\centering
\includegraphics[width=0.49\textwidth]{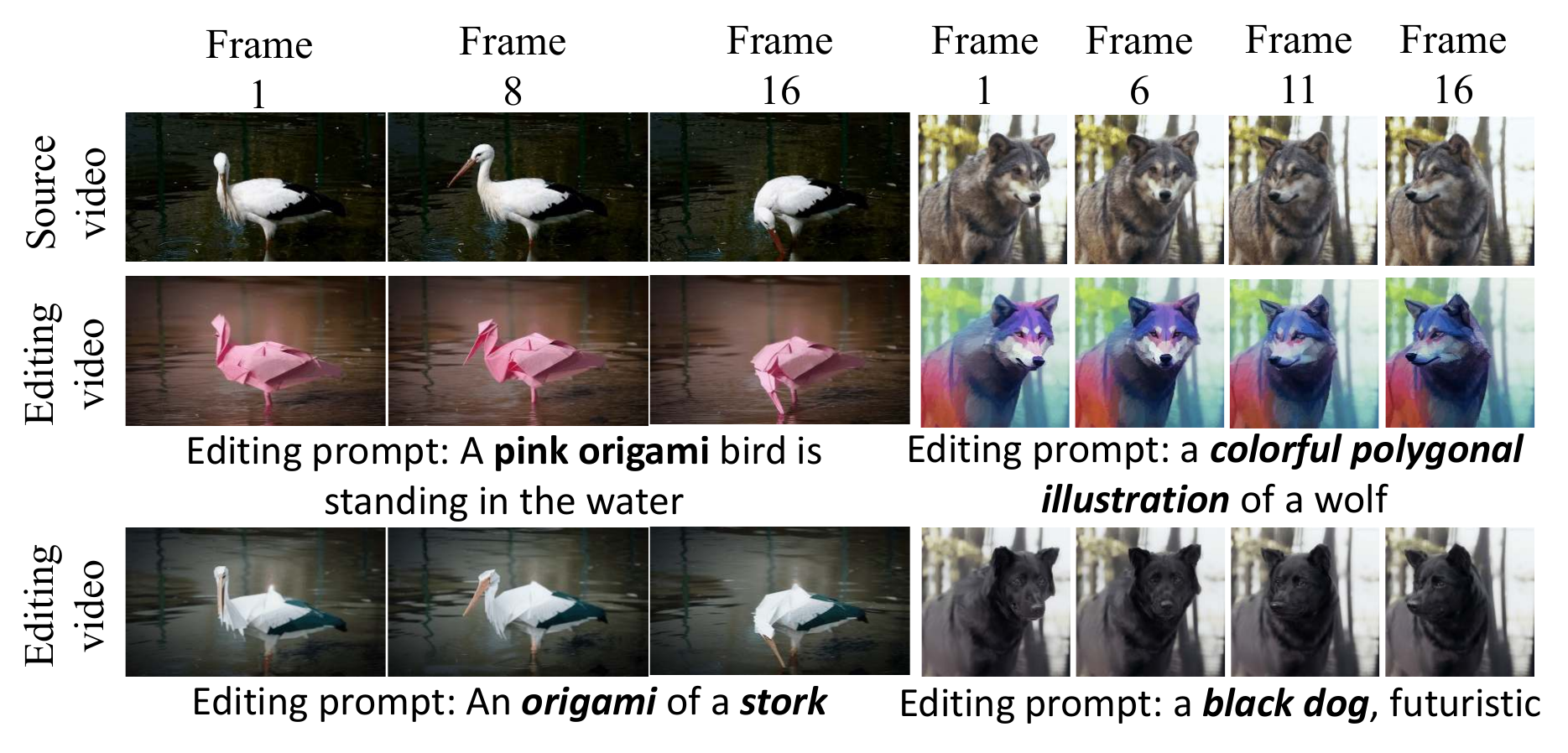}
\vspace{-1em}
\caption{Qualitative analysis of video editing on real-world videos.}
\vspace{-.5em}
\label{fig:real_edit}
\end{figure}

\begin{figure}[ht]
\centering
\includegraphics[width=0.49\textwidth]{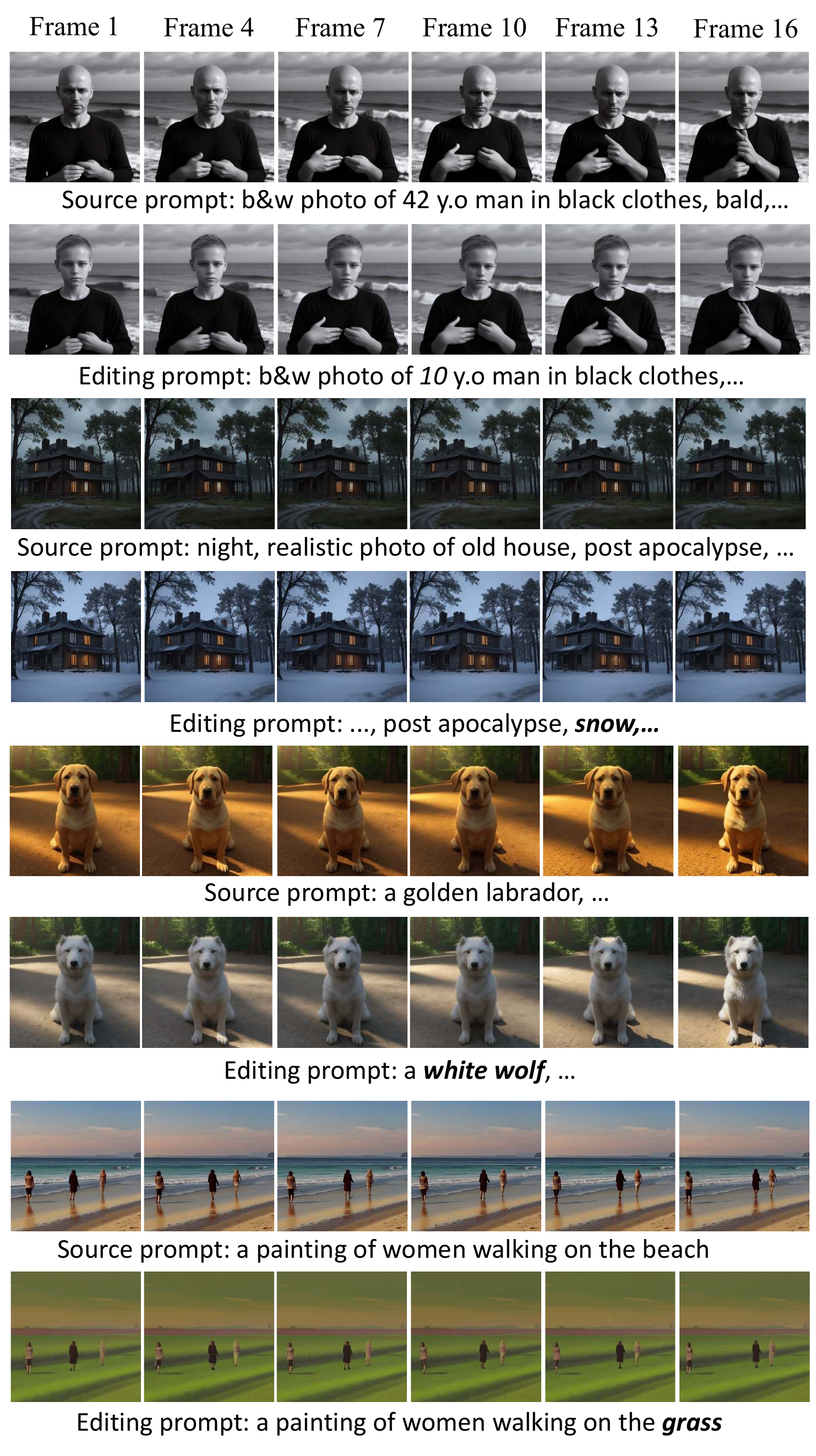}
\vspace{-1em}
\caption{Qualitative analysis of video editing on video-text pairs.}
\label{fig:video_editing_results}
\vspace{-1.5em}
\end{figure}

\begin{figure}[ht]
\centering
\includegraphics[width=0.490\textwidth]{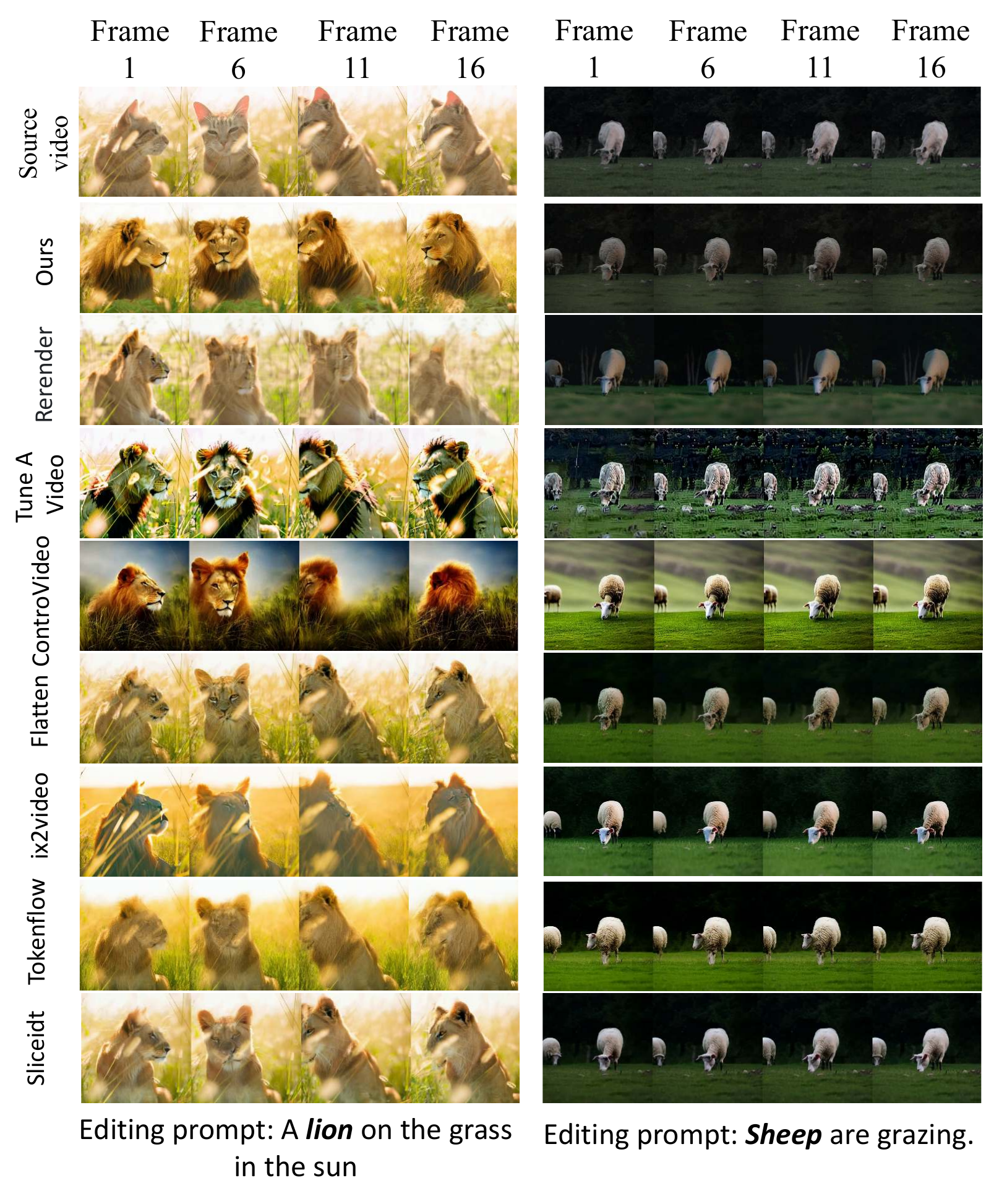}
\caption{Comparison of video editing results. From top to bottom: Ours, Rerender~\cite{yang2023rerender}, Tune-a-Video~\cite{wu2023tune}, ControlVideo~\cite{zhang2023controlvideo}, Flatten~\cite{flatten}, Pix2Video~\cite{Ceylan}, TokenFlow~\cite{tokenflow}, and Slicedit~\cite{slicedit}. Results of baselines are taken from the Slicedit paper~\cite{slicedit}. Note that only our method performs editing directly with a pure T2V model.}
\label{fig:editing_comparies}
\vspace{-1em}
\end{figure}

\subsection{Results of Video Editing}

We evaluate our method through both quantitative and qualitative analyses. As illustrated in Figures~\ref{fig:real_edit} and~\ref{fig:video_editing_results}, we present the editing results demonstrating its effectiveness in modifying various object attributes, video styles, scenes, and categories of the original video across multiple frames.

\noindent\textbf{Comparison to other video editing methods.}
We compare our work with state-of-the-art video editing methods, including (i) Rerender~\cite{yang2023rerender}, (ii) Tune-a-Video~\cite{wu2023tune}, (iii) ControlVideo~\cite{zhang2023controlvideo}, (iv) Flatten~\cite{flatten}, (v) Pix2Video~\cite{Ceylan}, (vi) TokenFlow~\cite{tokenflow}, and (vii) Slicedit~\cite{slicedit}. Figure~\ref{fig:editing_comparies} presents the comparative editing results. Notably, our method is unique in utilizing a pure T2V model, highlighting the potential of T2V frameworks in real-world video editing. Among the methods considered, Slicedit performs the best; however, it operates on a frame-by-frame basis using Stable Diffusion 2.1 and requires additional reference information for practical video editing.

\begin{figure}[ht]
\centering
\includegraphics[width=0.49\textwidth]{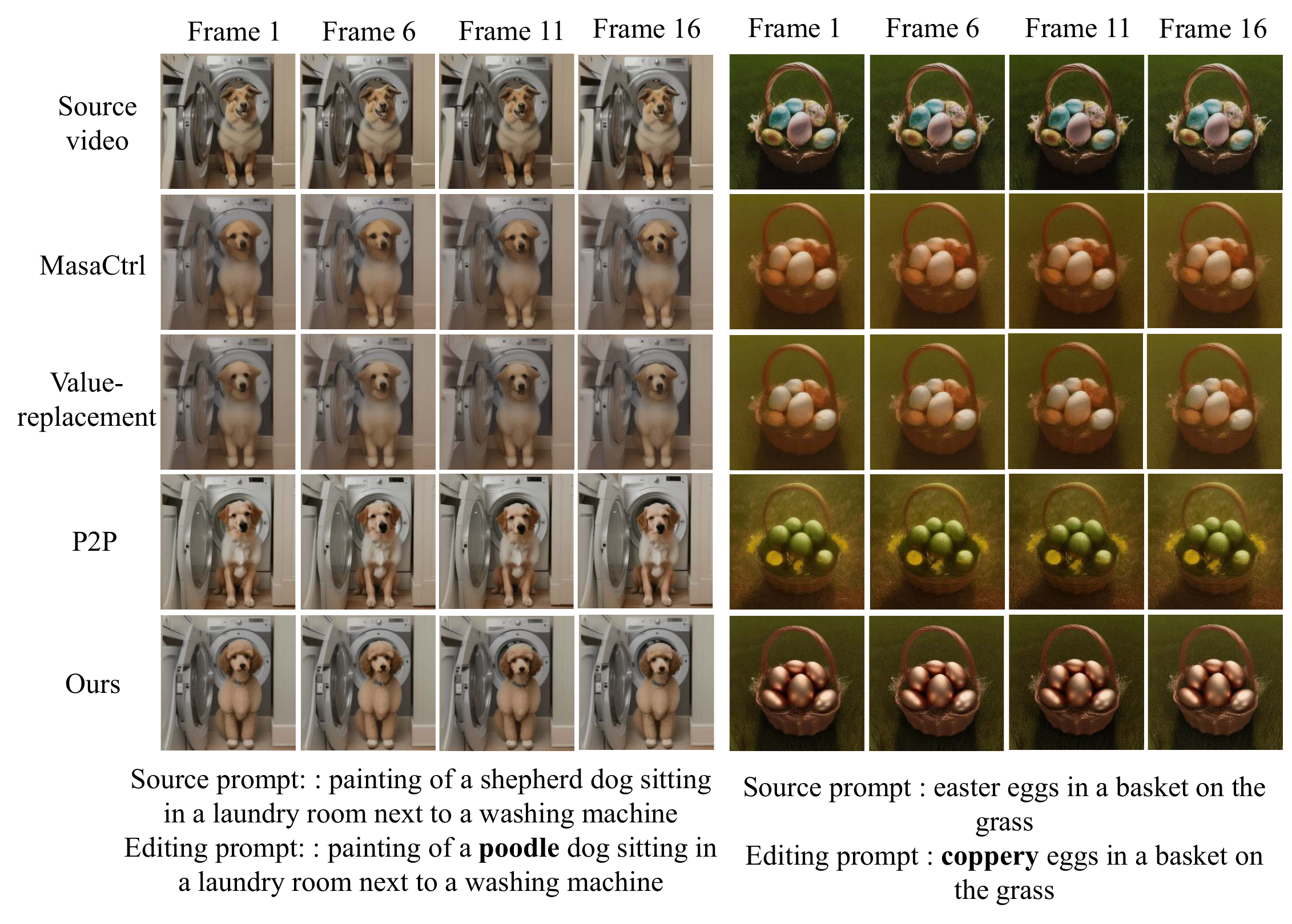}
\vspace{-1em}
\caption{Comparison results of different attention mechanism interventions on AnimateDiff, including Key-Value replacement (MasaCtrl)~\cite{masactrl}, Value replacement, and P2P~\cite{hertz2022prompt}.}
\label{fig:Diff_inject_method}
\vspace{-1em}
\end{figure}

Table~\ref{tab:editing_result} presents our quantitative results compared to other layer-to-editing strategies and attention replacement methods. Previous methods, such as TokenFlow~\cite{tokenflow}, empirically define attention layers for editing. 
MasaCtrl~\cite{masactrl} proposed to replace the \emph{Key} and \emph{Value} matrix in spatilal attention layer for image editing, while P2P~\cite{hertz2022prompt} proposed to replace the cross-attention map.
The comparative experiments of our method against various attention manipulation techniques are presented in Figure~\ref{fig:Diff_inject_method}.
Our approach specifies different editing layers for various editing instances. In the quantitative experimental results, our IE-Adapt strategy significantly outperforms the other two layer-to-editing strategies 
and various attention replacement methods in terms of text-image similarity, editing evaluation indicators, and inter-frame consistency, reflecting the effectiveness and value of IE-Adapt.

\begin{figure}[ht]
\centering
\begin{subfigure}{0.495\textwidth}
    \includegraphics[width=\textwidth]{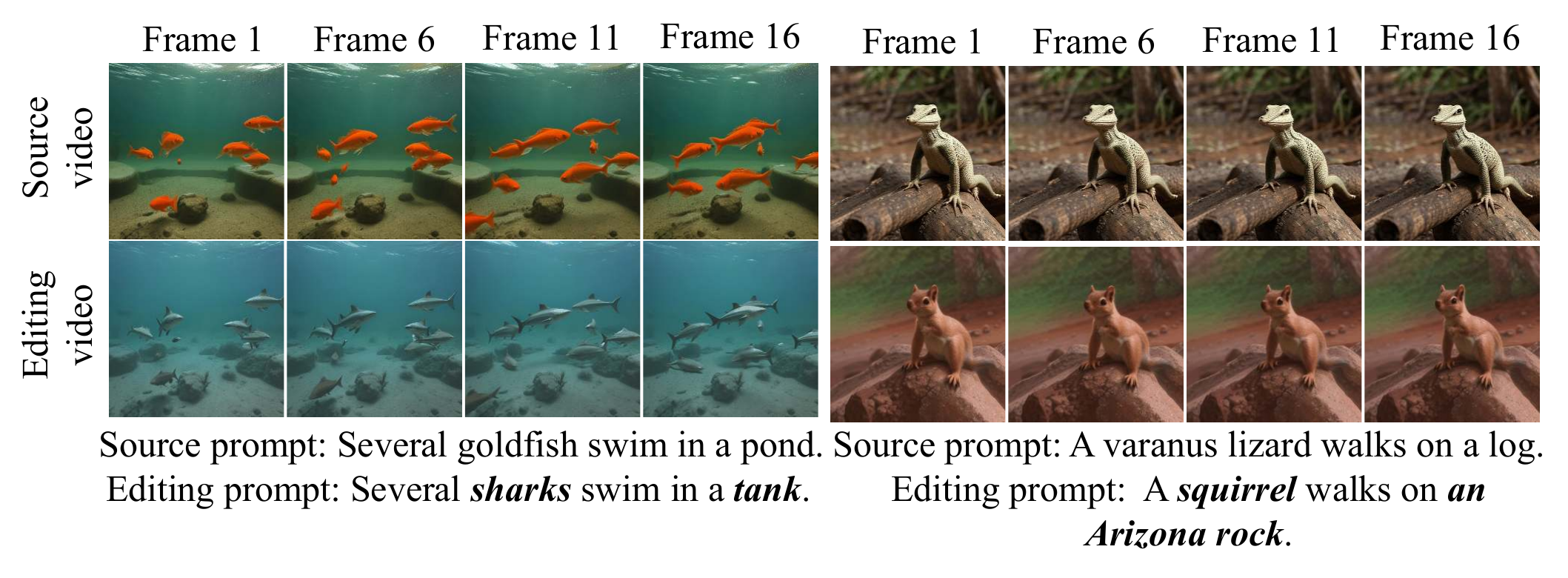}
    \caption{Video editing results with AnimateDiff-Lightning.}
    \label{fig:adlightning_results}
\end{subfigure}
\begin{subfigure}{0.495\textwidth}
    \includegraphics[width=\textwidth]{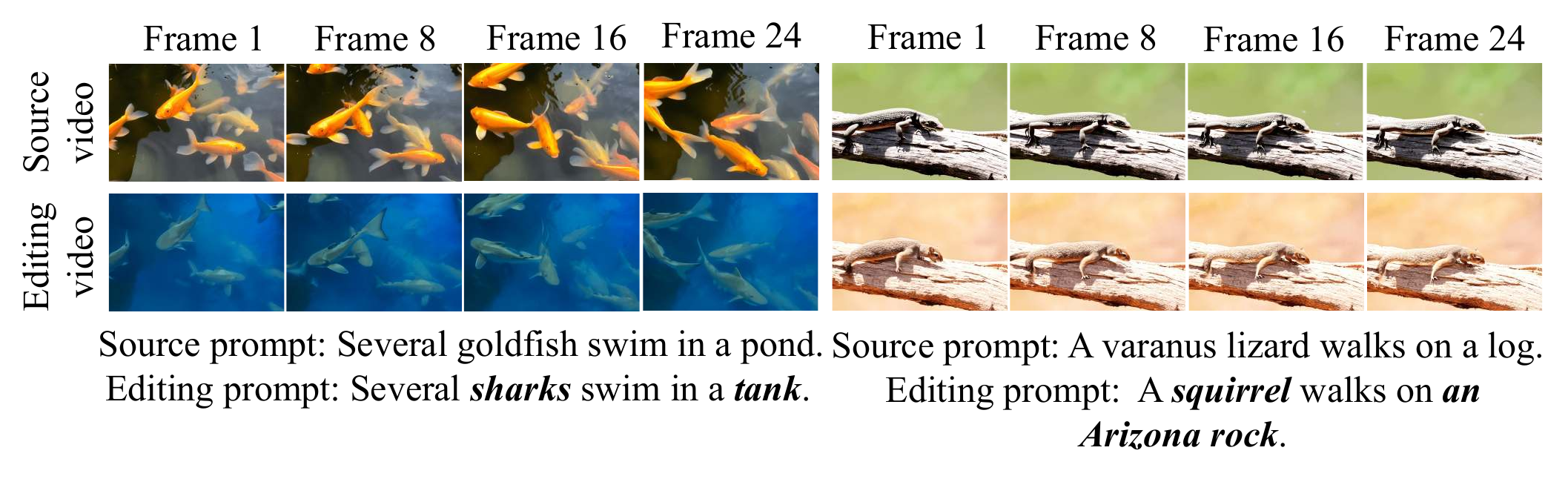}
    \caption{Video editing results with CogVideo5B.}
    \label{fig:cogvideo_results}
\end{subfigure}
\begin{subfigure}{0.495\textwidth}
    \includegraphics[width=\textwidth]{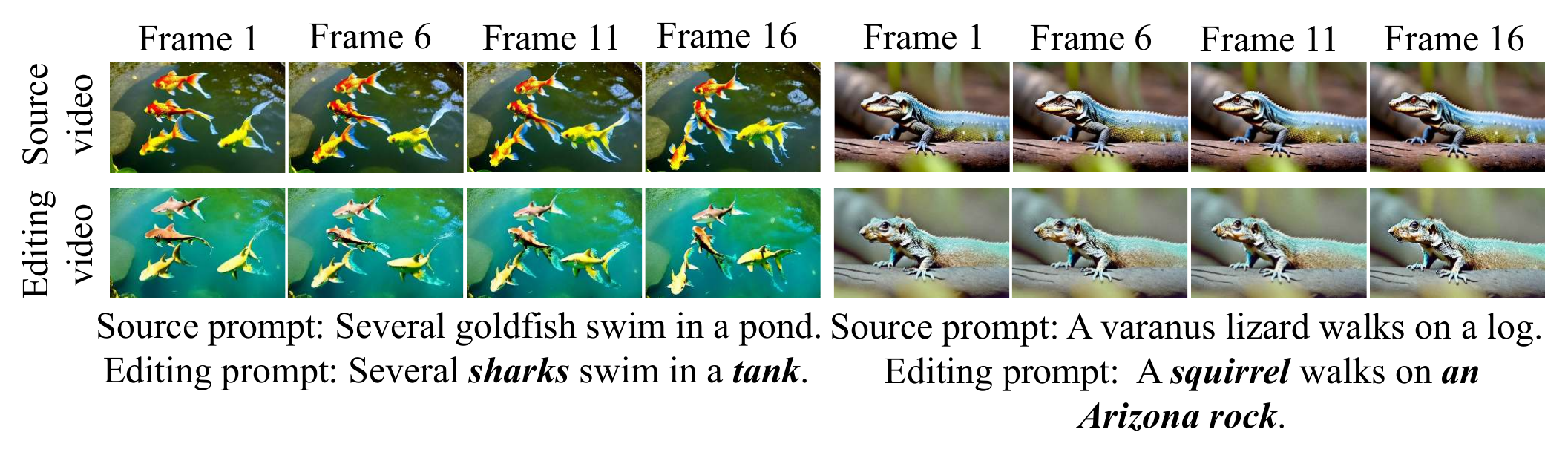}
    \caption{Video editing results with VideoCrafter2.}
    \label{fig:videocrafter2_edit}
\end{subfigure}
\vspace{-1em}
\caption{Video editing results of our method using other T2V models, including AnimateDiff-Lightning~\cite{AnimateDiff-Lightning}, CogVideoX-5B~\cite{Cogvideo5B} and VideoCrafter2~\cite{videocrafter2}.}
\label{fig:diff_model_results}
\vspace{-.5em}
\end{figure}

\noindent\textbf{Video editing using other VDMs.}
We further present the video editing results using other VDMs to demonstrate the universality of our method, including AnimateDiff-Lightning~\cite{AnimateDiff-Lightning}, CogVideoX-5B~\cite{Cogvideo5B}, and VideoCrafter2~\cite{videocrafter2}. The editing results are shown in Figure~\ref{fig:diff_model_results}. In each case, our method successfully edits the input videos according to the editing prompts with VDMs, demonstrating the effectiveness and adaptability of our approach. Please see the supplementary materials for more experimental results.

\begin{table}[ht]
\centering
\aboverulesep=0pt
\belowrulesep=0pt
\begin{footnotesize}
\begin{tabular}{c|cccc}
    \toprule
    Method & CS$\uparrow$  & CDS $\uparrow$ & MS $\uparrow$ & SC $\uparrow$ \\
    \hline
    All layers& 27.94 & 0.1888 & 0.9766 & \underline{0.9946} \\
    \hline
    TokenFlow$^{\ast}$& 27.78 & 0.1830 & 0.9768 & \textbf{0.9956}\\
    \bottomrule
    MasaCtrl$^{\ast}$ & 27.67 & 0.1228  & 0.9748  &0.9927   \\
    Value & 27.68 & 0.1245 &0.9750& 0.9929  \\
    P2P$^{\ast}$ & \underline{28.45}& \underline{0.2048} &  \textbf{0.9796} & 0.9908  \\
    \midrule
    Ours & \textbf{29.38} & \textbf{0.2475} & \underline{0.9781} & 0.9913 \\
    \bottomrule
\end{tabular}
\end{footnotesize}
\caption{Quantitative results of video editing on AnimateDiff. TokenFlow$^{\ast}$ refers to a class of methods \cite{tokenflow,ku2024anyv2v} that perform editing operations on attention maps from the 4th to 11th layers of decoder blocks. MasaCtrl$^{\ast}$ (Key-Value replacement) \cite{masactrl}, Value replacement, and P2P$^{\ast}$ \cite{hertz2022prompt} represent different replacement methods.}
\label{tab:editing_result}
\vspace{-1.25em}
\end{table}

\section{Conclusion}
In conclusion, our study provides new insights into the role of attention in diffusion-based T2V models. By conducting a perturbation analysis, we show the significant influence of attention maps' entropy on video quality, demonstrating the importance of these components in controlling video dynamics and structure. Building on these findings, we introduce two lightweight methods that enhance video quality and facilitate text-guided video editing. Integrating these methods into existing T2V workflows can significantly improve video generation and editing efficiency. Our results demonstrate the robustness and effectiveness of these approaches. In the future, we plan to extend our findings to other applications based on T2V models.

\newpage

{
    \small
    \bibliographystyle{ieeenat_fullname}
    \bibliography{main}
}

\clearpage
\setcounter{page}{1}
\maketitlesupplementary
The Supplementary Material of this paper consists of several sections that provide additional information and support for the main content.
\begin{itemize}
    \item \textbf{Perturbation Analysis Results:} This section presents the complete results of the perturbation experiments, including detailed perturbation results for the AnimateDiff and CogVideoX models.
    \item \textbf{Algorithm for Video Editing:} This section presents the algorithm for our video editing approach.
    \item \textbf{Video Generation Quality Improvement Results:} This section provides additional experimental results for our video generation improvement approach.
    \item \textbf{Video Editing Results:} This section offers further experimental results for our video editing approach.
\end{itemize}

\section{Results of Perturbation Analysis } 
\subsection{Implementation Details}
Our experimental setup is described as follows: For video generation using AnimateDiff,\footnote{T2I-based model: \url{https://huggingface.co/SG161222/Realistic_Vision_V5.1_noVAE}\\ Motion adapter model: \url{https://huggingface.co/guoyww/animatediff-motion-adapter-v1-5-3}} the model is initialized with a `torch.float16` data type. Each video consists of 16 frames, each with a resolution of \(512 \times 512\). We use a random seed of 42, set the guidance scale to 9.0, conduct 25 inference steps, and utilize the Euler Scheduler for scheduling. For the CogVideoX-5B model,\footnote{\url{https://huggingface.co/THUDM/CogVideoX-5b}} the generated videos comprise 25 frames, each with a resolution of \(480 \times 720\). Similarly, the random seed is 42, the guidance scale is 6.0, and the number of inference steps is 25.

\subsection{Perturbation Results on AnimateDiff}

In this section, we further supplement the content of Section~\ref{pert_atten_map} in the main text of the paper.

\begin{figure*}[ht]
    \includegraphics[width=0.90\textwidth]{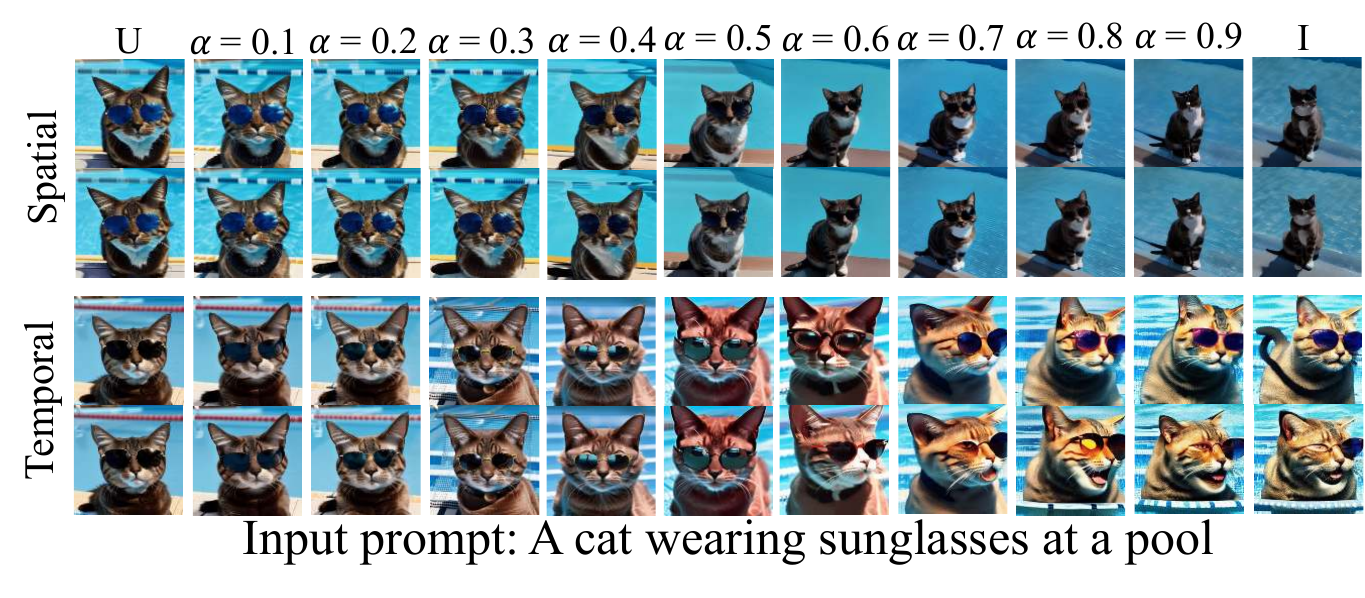}
    \caption{Perturbation results using the attention map between \(U\) and \(I\).}
    \label{fig:betweenUI}
\end{figure*}

\noindent\textbf{Perturbation analysis using a matrix between \(U\) and \(I\).} 
We also introduced a new method for constructing attention maps: \(\alpha \times I + (1-\alpha) \times U\), perturbing the attention between \(U\) and \(I\). Figure~\ref{fig:betweenUI} illustrates a gradual change in results.

\begin{figure*}[ht]
    \includegraphics[width=0.90\textwidth]{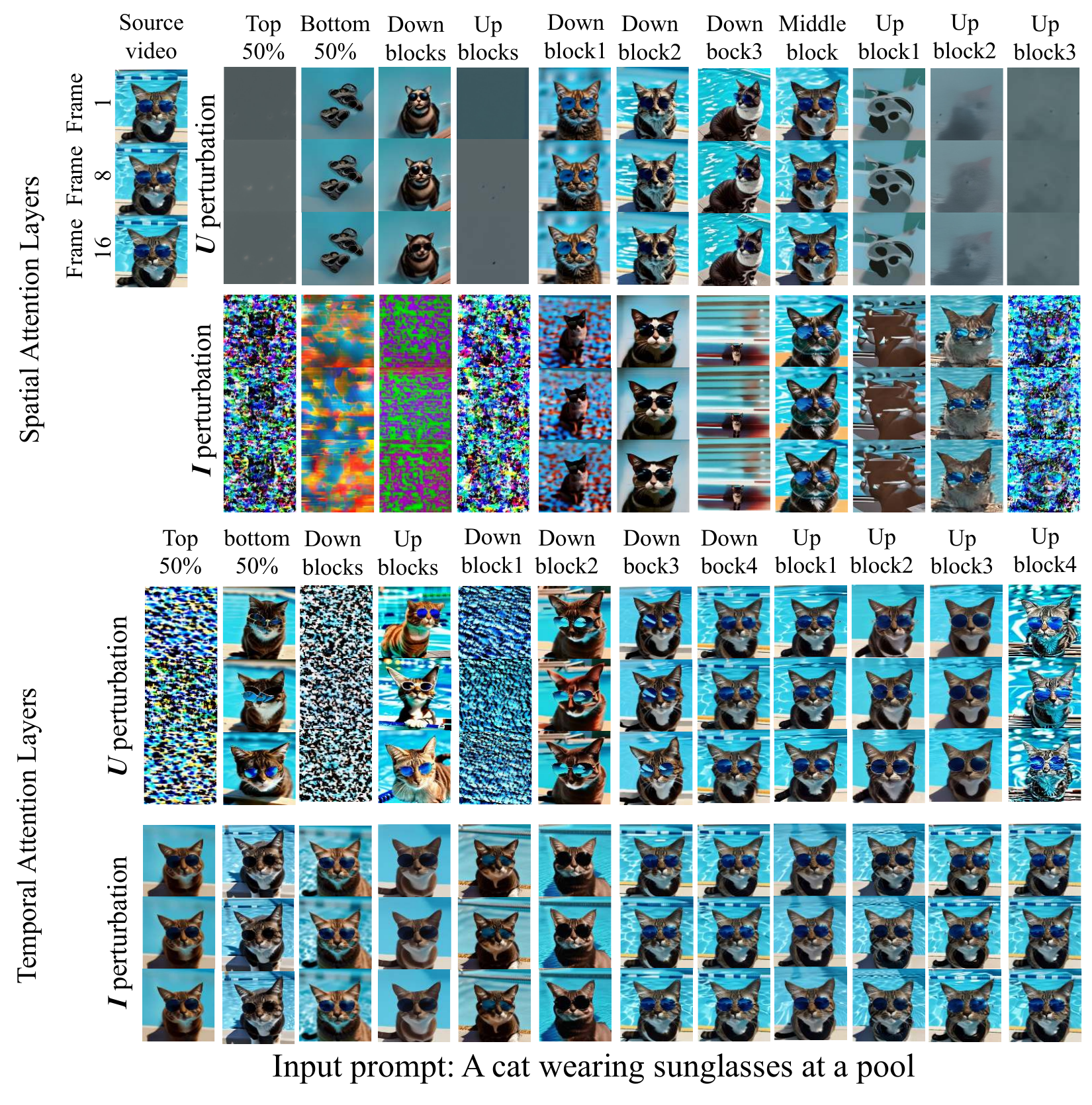}
    \caption{Perturbation results beyond a single layer.}
    \label{fig:diff_layers}
\end{figure*}

\noindent\textbf{Perturbation analysis beyond a single layer.}
To minimize variability in perturbation analysis, we focused on a single layer. Figure~\ref{fig:diff_layers} displays results for various layer combinations. 
We conducted experiments with multi-layer perturbations using the following combinations: the 50\% layers with the highest and lowest entropy, the upsampling and downsampling layers of the attention layer, and the submodules in the upsampling and downsampling layers, as shown in Figure~\ref{fig:diff_layers}. Perturbing multiple layers often leads to significant result changes. Comparatively, \(U\) perturbations at the block level of the temporal layer have a smaller impact, consistent with the results of perturbing single layers. The initial and final layers of the U-Net network are crucial for establishing intra-frame quality and inter-frame relationships.

\begin{figure*}[ht]
\centering
\includegraphics[width=0.85\textwidth]{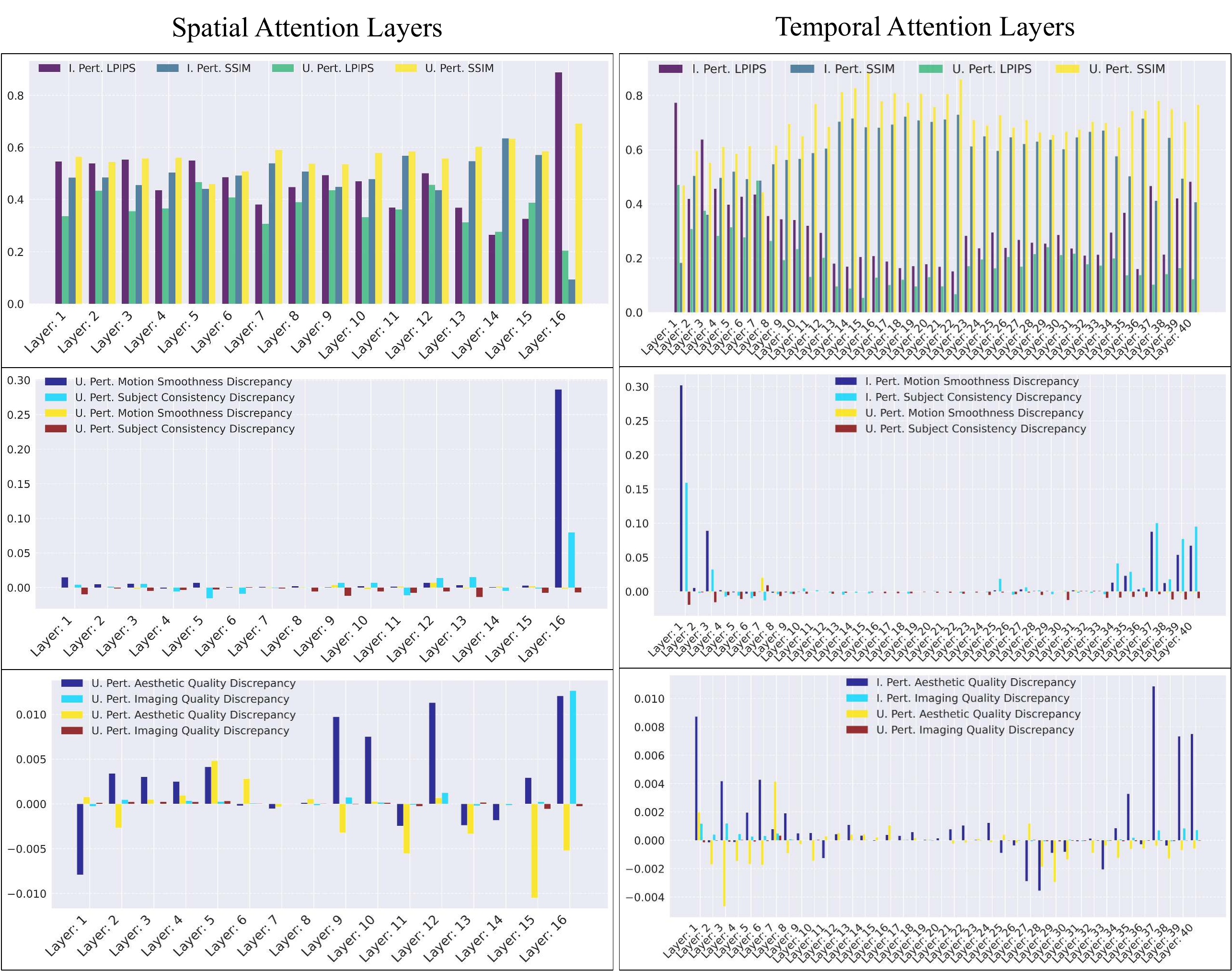}
\caption{Histogram of attention map perturbation results on AnimateDiff. From top to bottom: structural measure, temporal consistency difference before and after perturbation, and shift in aesthetics. Smaller values for LPIPS indicate better performance, while larger values are preferred for the other metrics.}
\label{fig:suppl_AD_h}
\vspace{-1em}
\end{figure*}

\begin{figure*}[ht]
\centering
\includegraphics[width=0.85\textwidth]{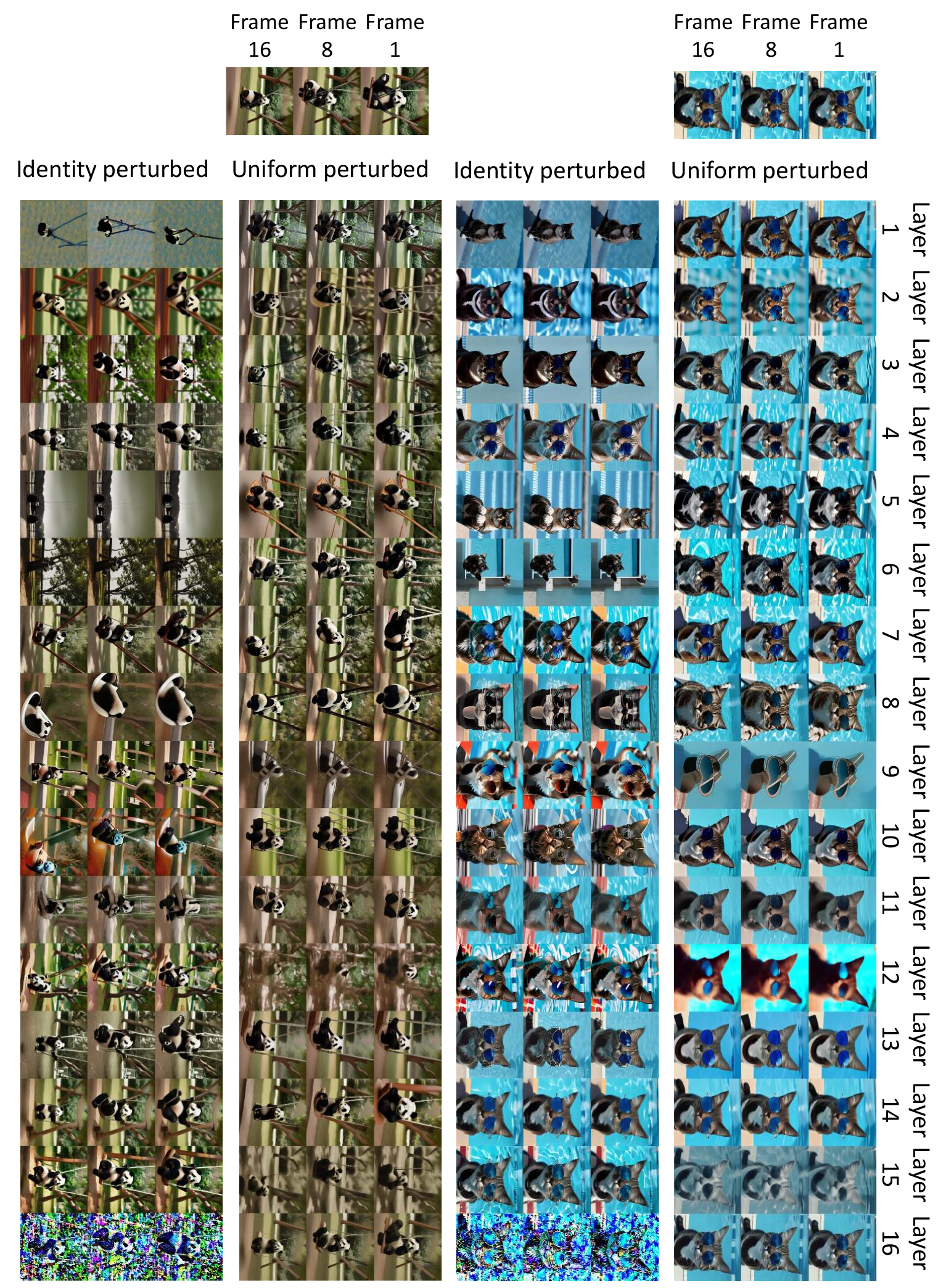}
\caption{Perturbation results on AnimateDiff. $I$ perturbation refers to replacing $A$ with $I$ at the $L$-th layer, while $U$ perturbation substitutes it with a uniform matrix. The input prompts are selected from VBench. Perturbation is conducted on both spatial and temporal attention layers, where only one layer is perturbed at a time.}
\label{fig:suppl_pert_spatial_ad}
\vspace{-1em}
\end{figure*}

\begin{figure*}[ht]
\centering
\begin{subfigure}{0.75\textwidth}
    \includegraphics[width=\textwidth]{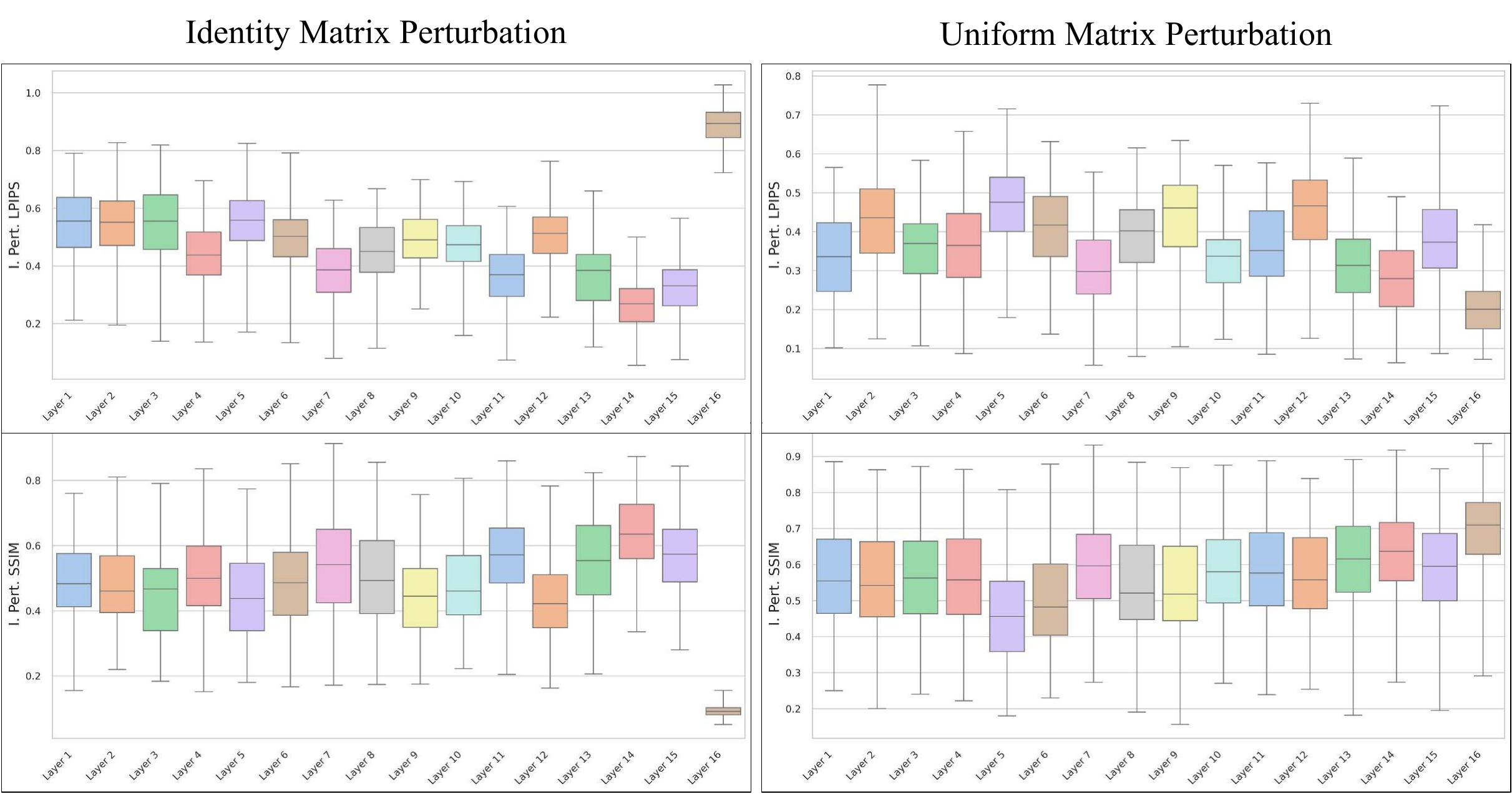}
    \caption{Box of attention map perturbation results on structural measure}
    \label{fig:AD_box_Spa_LPIPS_supp}
\end{subfigure}

\begin{subfigure}{0.75\textwidth}
    \includegraphics[width=\textwidth]{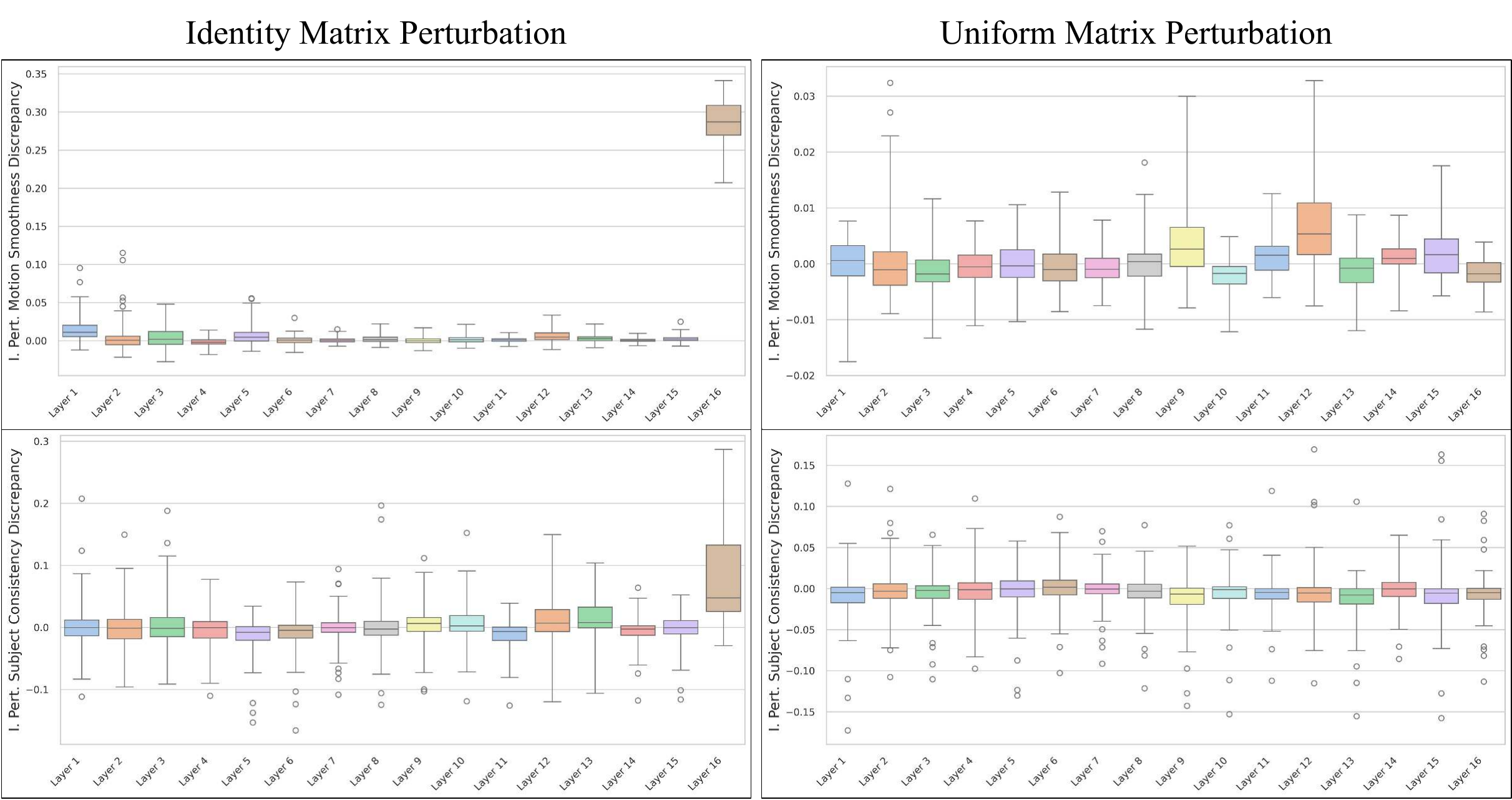}
    \caption{Box of attention map perturbation results on temporal consistency}
    \label{fig:AD_box_Spa_MS_supp}
\end{subfigure}
\begin{subfigure}{0.75\textwidth}
    \includegraphics[width=\textwidth]{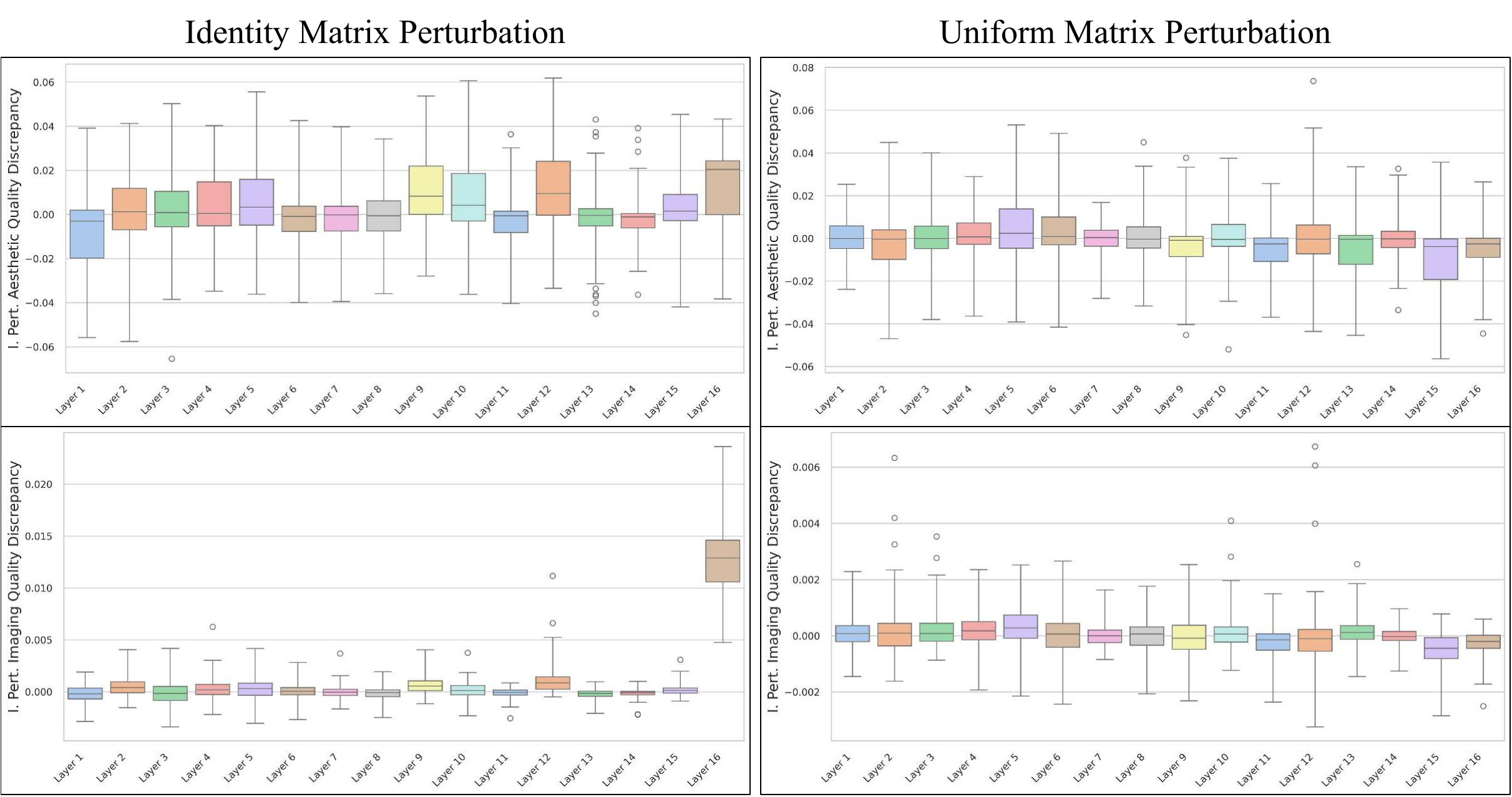}
    \caption{Box of attention map perturbation results on aesthetics shift}
    \label{fig:AD_box_Spa_AS_supp}
\end{subfigure}
\caption{Box of attention map perturbation results on spatial attention layers in AnimateDiff.}
\label{fig:suppl_AD_spatial_box}
\end{figure*}

\begin{figure*}[ht]
\centering
\begin{subfigure}{0.75\textwidth}
    \includegraphics[width=\textwidth]{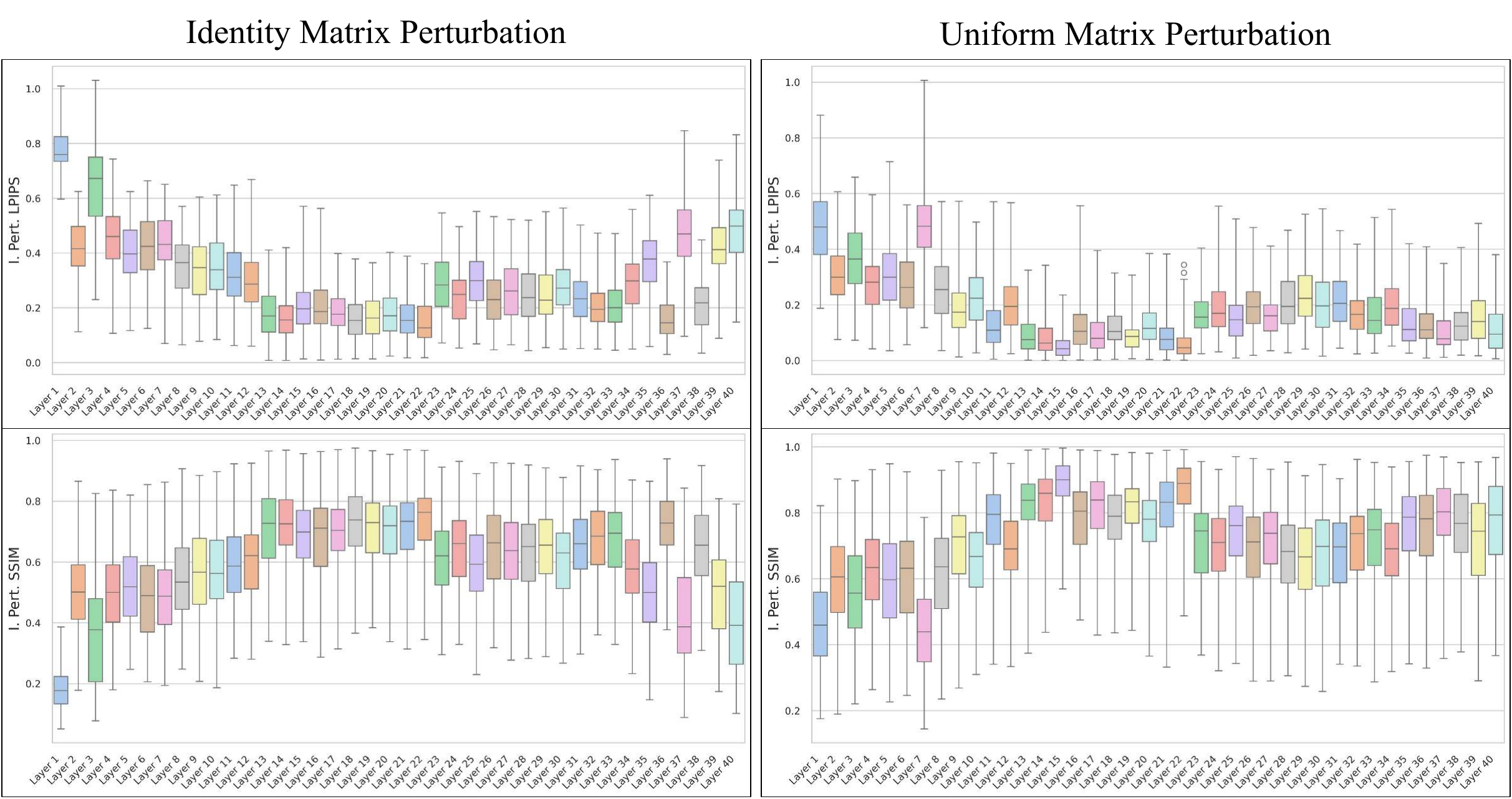}
    \caption{Box of attention map perturbation results on structural measure}
    \label{fig:AD_box_Temp_LPIPS_supp}
\end{subfigure}

\begin{subfigure}{0.75\textwidth}
    \includegraphics[width=\textwidth]{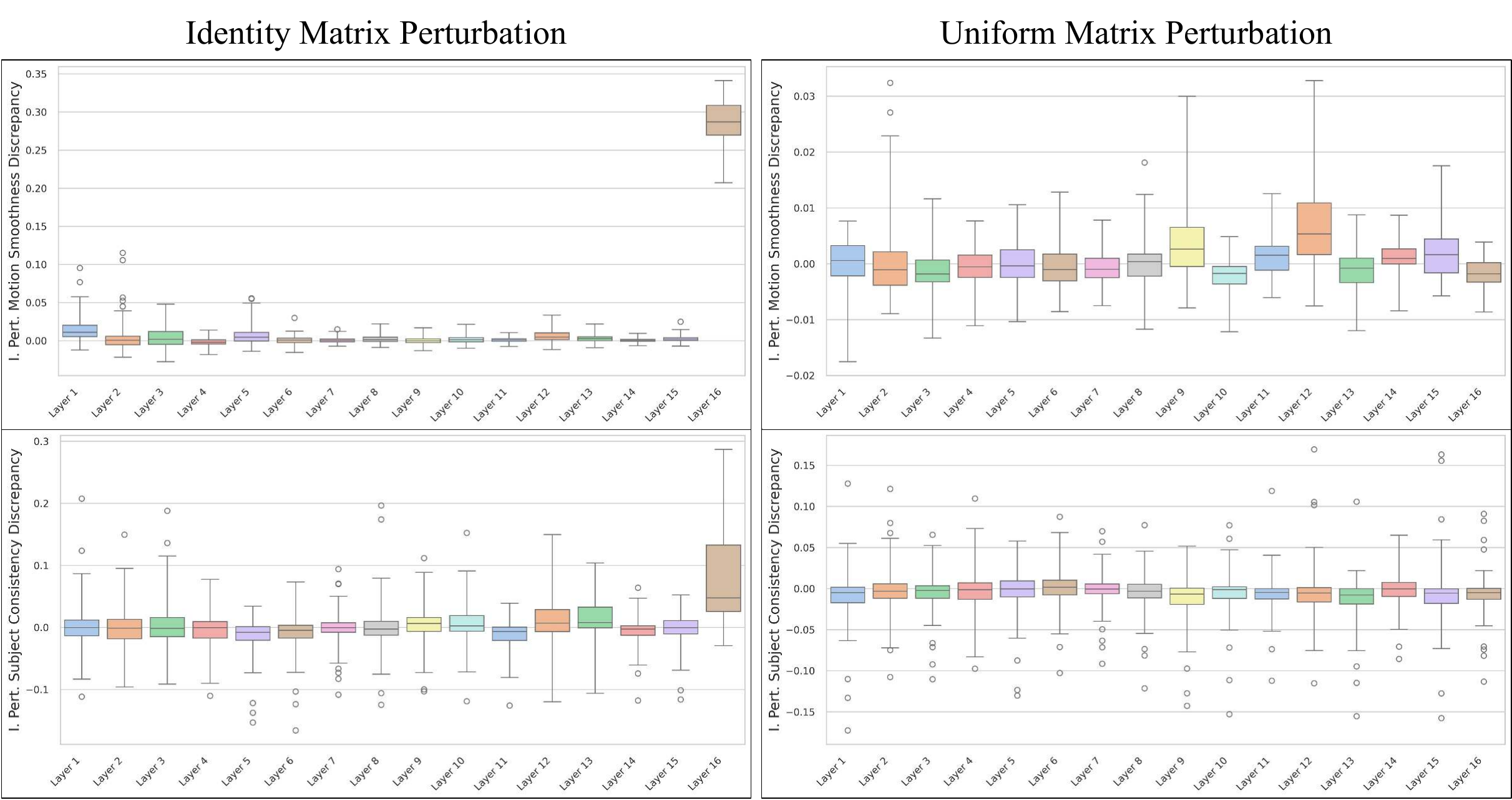}
    \caption{Box of attention map perturbation results on temporal consistency}
    \label{fig:AD_box_Temp_MS_supp}
\end{subfigure}
\begin{subfigure}{0.75\textwidth}
    \includegraphics[width=\textwidth]{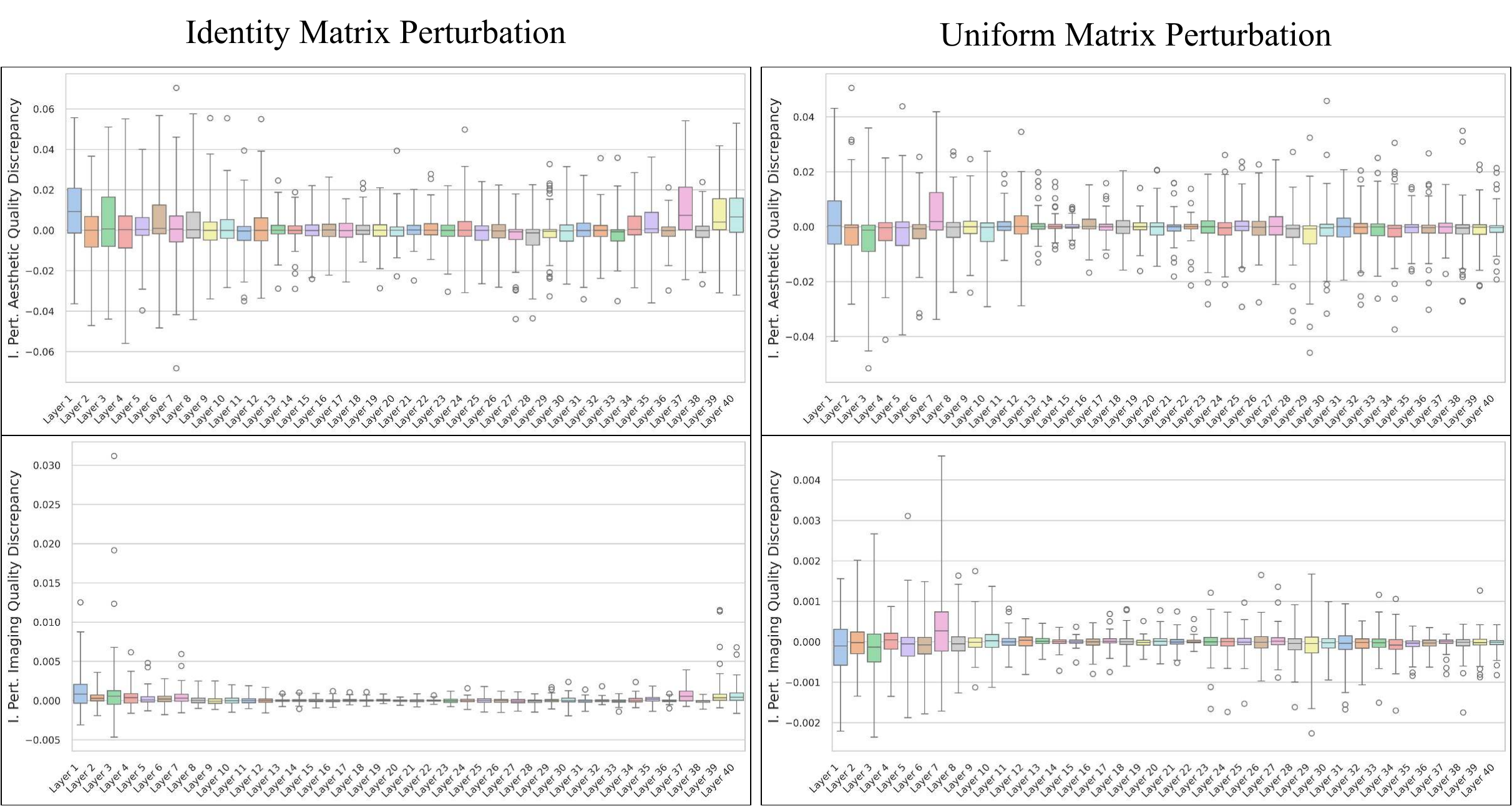}
    \caption{Box of attention map perturbation results on aesthetics shift}
    \label{fig:AD_box_Temp_AS_supp}
\end{subfigure}
\caption{Box of attention map perturbation results on temporal attention layers in AnimateDiff.}
\label{fig:suppl_AD_temporal_box}
\end{figure*}

\noindent\textbf{Complete supplement to the perturbation analysis in the main text.}
Figure~\ref{fig:suppl_pert_spatial_ad} is a complete supplement to Figure~\ref{fig:perturbed_video_case}, and Figure~\ref{fig:suppl_AD_h} is a complete supplement to Figure~\ref{fig:Perturbed_histogram}. Additionally, Figure~\ref{fig:suppl_AD_spatial_box} and Figure~\ref{fig:suppl_AD_temporal_box} show the box plots of the distribution of different evaluation indicators after the perturbation of different attention layers.

From the perspective of perturbation, both the identity matrix and the uniform matrix affect the structure of the source video at each layer. However, compared with the uniform matrix, the identity matrix has a more significant effect on the structure. At the temporal layer, the uniform matrix has little impact on the structure within the video frame. The identity matrix perturbation can destroy the consistency between video frames and affect the imaging quality within the video frame. In contrast, the uniform matrix perturbation shows a limited effect on frame consistency, although perturbation on some layers can also improve the aesthetics within the video frame.

The box plot can reflect the degree of fluctuation of the perturbation of different samples on the same layer. In the perturbation experiments concerning frame structure, the evaluation indicators for all samples across different layers exhibit minimal fluctuation, with no outliers observed. In terms of imaging quality and intra-frame consistency, fluctuations remain small; however, a limited number of outliers are present. Generally, the trend of the median line in the box plot is consistent with the trend of the histogram in Figure~\ref{fig:Perturbed_histogram}.

\subsection{Perturbation Results on CogVideoX}

\begin{figure*}[ht]
\centering
\includegraphics[width=0.80\textwidth]{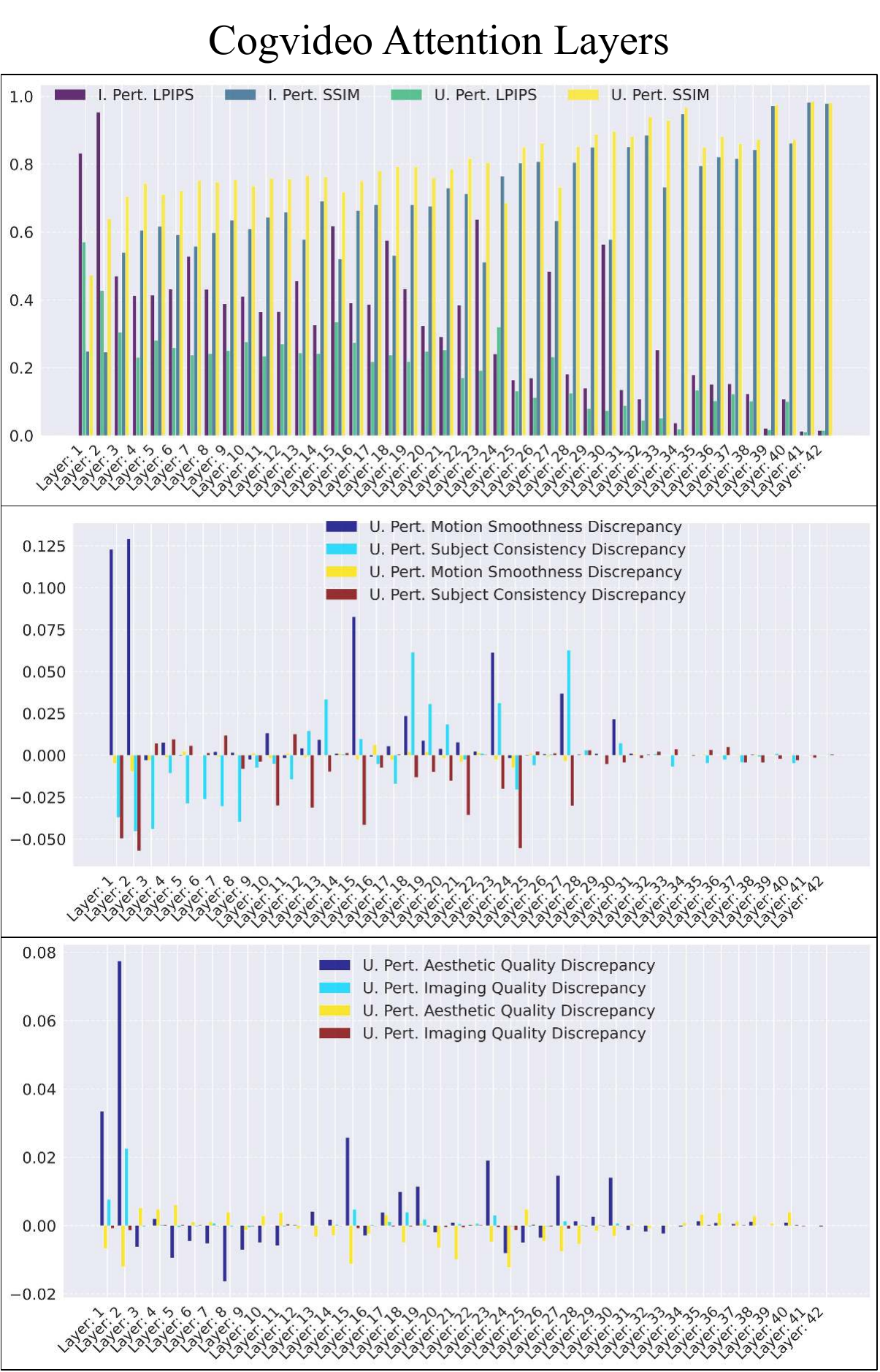}
\caption{Histogram of attention map perturbation results on CogVideoX. From top to bottom: structural measure, temporal consistency difference before and after perturbation, and shift in aesthetics. Smaller values for LPIPS indicate better performance, while larger values are preferred for the other metrics.}
\label{fig:suppl_cogvideo_h}
\end{figure*}

For CogVideoX, the initial attention layer plays a key role in determining the quality of the generated video, whether it involves the construction of a spatial structure, the maintenance of inter-frame relations, or the improvement of intra-frame image quality. Shallow attention layers are susceptible to perturbations. When perturbations are applied to these attention maps, the intra-frame structure, aesthetics, and inter-frame consistency of the video are significantly impacted.
The evaluation indicators in Figure~\ref{fig:suppl_cogvideo_h} reveal the impact of perturbations. Specifically, replacing the initial attention layer with \(I\) or \(U\) significantly alters the intra-frame image structure of the video. However, replacing the attention map in a similar way in the deep layers of the model does not have a significant effect. As shown in Figure~\ref{fig:cogvideo_entropy}, the information entropy of the attention maps of all layers in CogVideoX ranges from 15\% to 45\%. This shows that the model's attention map is closer to the identity matrix than the uniform matrix; that is, it captures more local structural information and relational information of adjacent frames. This is significantly different from AnimateDiff. Therefore, replacing these attention maps with \(I\) or \(U\) will cause drastic changes in the output energy. This modification may destroy the structural features of the original image, resulting in a change in image structure, which can be reflected by lower SSIM and higher LPIPS scores.
In addition, perturbations to the deep attention maps have almost no effect on the video results. We speculate that this is because these layers contribute less to the network information flow, with more information coming from the tail skip connection. In subsequent research, we will conduct an in-depth analysis of this phenomenon.
Overall, CogVideoX's shallow attention maps are crucial to the final generated video result, so it is not recommended to modify them in the shallow layers to improve video quality or perform video editing. In contrast, the deep attention maps have less impact on the output results, and intervening in them may not be beneficial. We recommend intervening in the middle layers with smaller information entropy to achieve better results in improving video quality and editing effects.

\begin{figure*}[ht]
\centering
\begin{subfigure}{0.75\textwidth}
    \includegraphics[width=\textwidth]{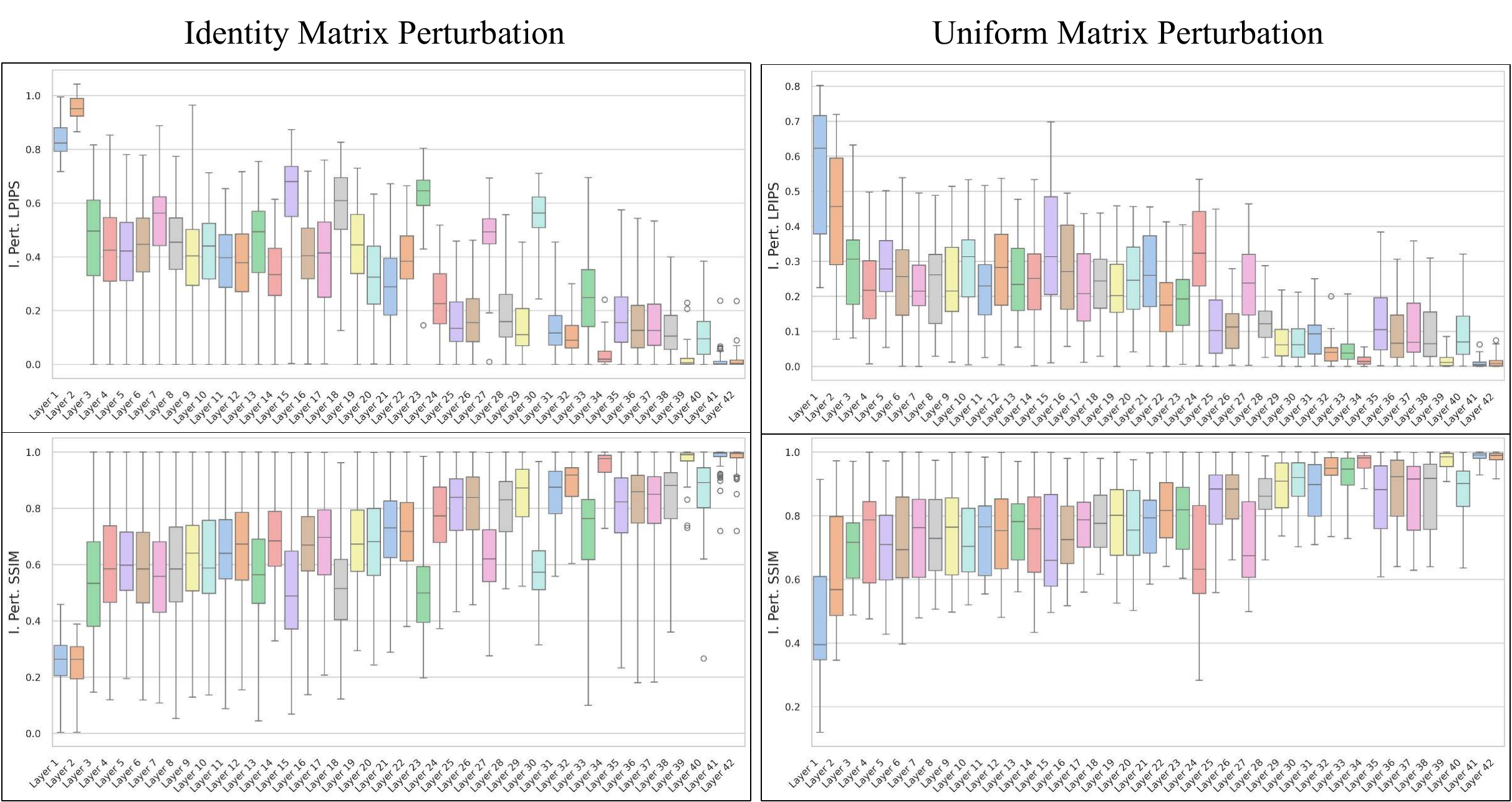}
    \caption{Box of attention map perturbation results on structural measure}
    \label{fig:cogvideo_box_LPIPS_supp}
\end{subfigure}

\begin{subfigure}{0.75\textwidth}
    \includegraphics[width=\textwidth]{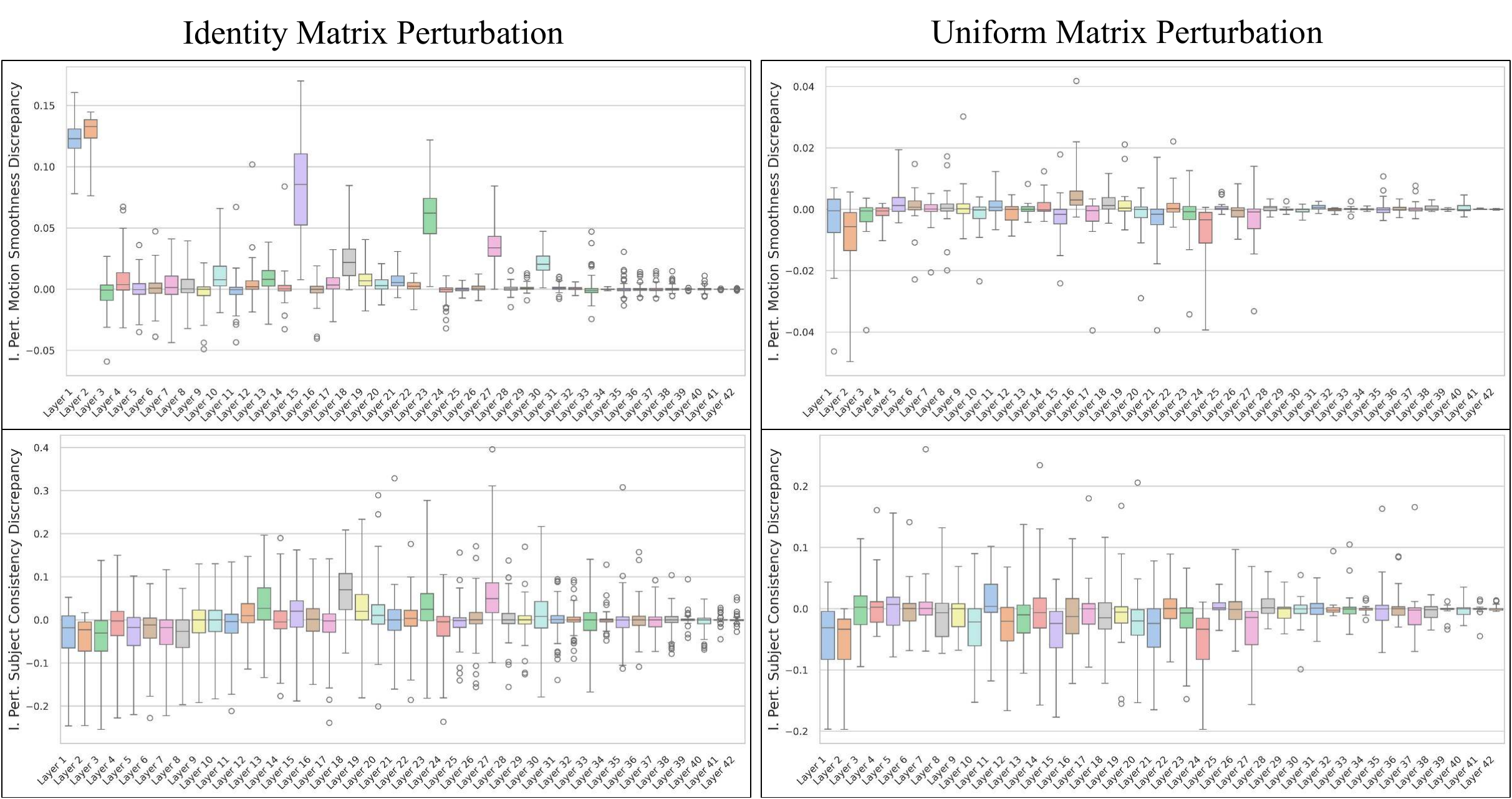}
    \caption{Box of attention map perturbation results on temporal consistency}
    \label{fig:cogvideo_box_MS_supp}
\end{subfigure}
\begin{subfigure}{0.75\textwidth}
    \includegraphics[width=\textwidth]{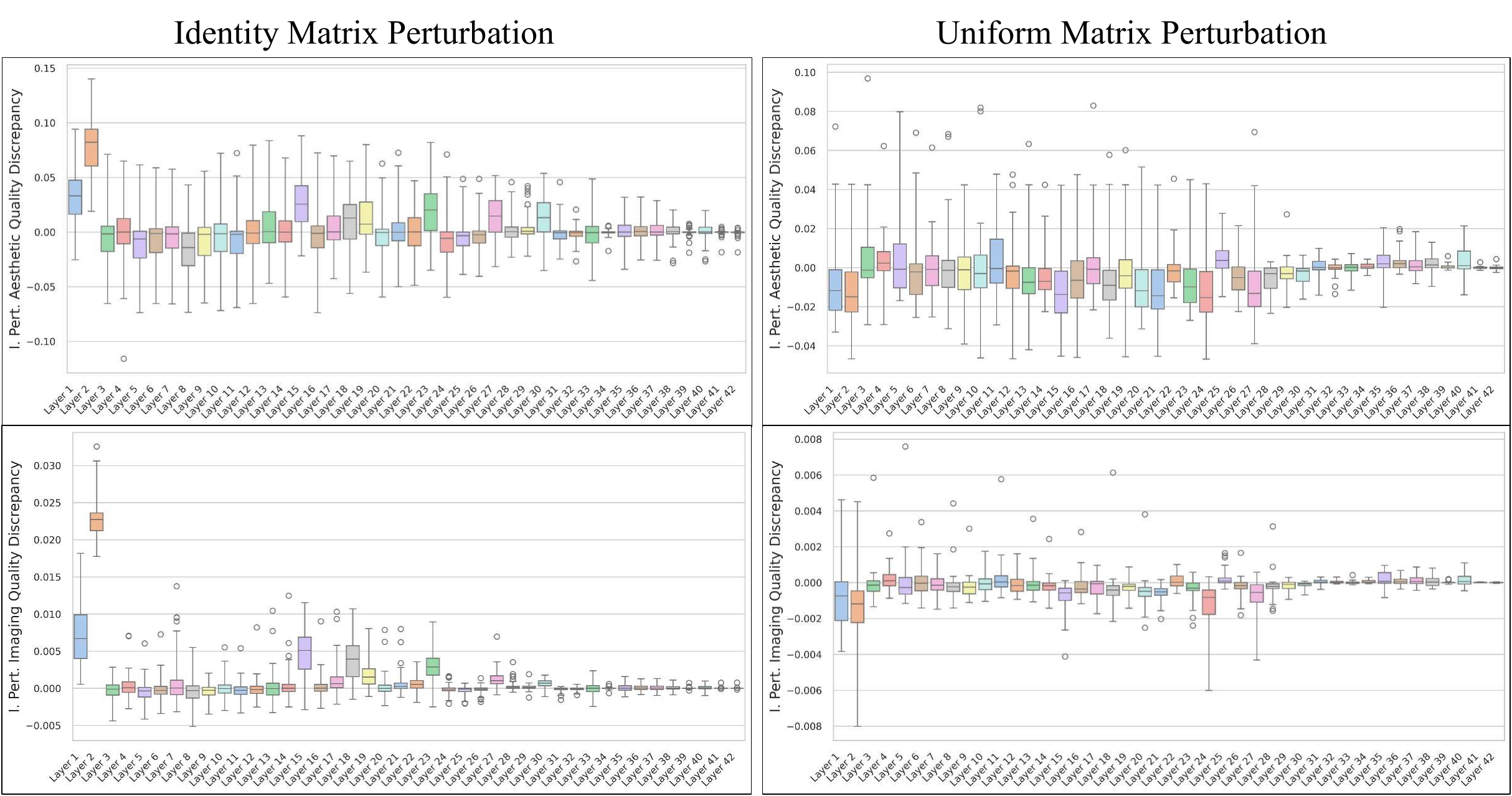}
    \caption{Box of attention map perturbation results on aesthetics shift.}
    \label{fig:cogvideo_box_AS_supp}
\end{subfigure}
\caption{Box of attention map perturbation results on CogVideoX.}
\label{fig:suppl_cogvideo}
\end{figure*}

\subsection{Video Editing Algorithm}

The algorithm of our video editing algorithm is shown below.

\begin{algorithm}[!ht]
\footnotesize
    \renewcommand{\algorithmicrequire}{\textbf{Input:}}
	\renewcommand{\algorithmicensure}{\textbf{Output:}}
	\caption{Editing Videos Generated by T2V Models}
    \label{Gen_algorithm}
    \begin{algorithmic}[1] 
        \REQUIRE  $P_{src}$: a source prompt;
                  $P_{dst}$: a target prompt;
                  $S$: random seed;
	    \ENSURE $V_{src}$: source video;
                $V_{dst}$: edited video;
        
        \STATE $z^{1:f}_{T} \sim \mathcal{N}(0,1)$, a unit Gaussian noise sampled with seed $S$;
        \STATE $A^{1:n} \leftarrow \mathcal{F}(z^{1:f}_{t}, P_{src})$;
        \STATE $\mathcal{I}\leftarrow$ indices of $A^{1:n}$ with the lowest 50\% of entropy;
        
        \STATE $\mathcal{F}^{*} \leftarrow \text{IE-Adapt}(\mathcal{F}, \mathcal{I})$;
        \STATE $\hat{z}^{1:f}_{T} \leftarrow z^{1:f}_{T}$;
        
        \FOR {$t = T, T-1, \ldots, 1$}
            \STATE $\epsilon^{1:f}_{t}, A \leftarrow \mathcal{F}^{*}(z^{1:f}_{t}, P_{src}, t)$;
            \STATE $\hat{\epsilon}^{1:f}_{t} \leftarrow \mathcal{F}^{*}(\hat{z}^{1:f}_{t}, P_{dst}, t); \hat{A} \leftarrow A$;
            \STATE $\hat{z}^{1:f}_{t-1} \sim \mathcal{N}\left(\frac{1}{\sqrt{\overline{\alpha}_t}} \left(x_t - \frac{1-\alpha_t}{\sqrt{1-\overline{\alpha}_t}}\hat{\epsilon}^{1:f}_{t}\right), \Sigma_t\right)$ \text{by Eq.~\ref{eq:x_sample}};
        \ENDFOR
        \STATE \textbf{Return} $(V_{src} \leftarrow \text{Decoder}(z^{1:f}_{0}), V_{dst} \leftarrow \text{Decoder}(\hat{z}^{1:f}_{0}))$;
    \end{algorithmic}
\end{algorithm}

\section{Experimental Results} 

\subsection{Results of Video Quality Enhancement}

We further discuss more improvement strategies using 93 evaluation prompts as follows:
\begin{equation}
\begin{split}
\text{Strategy 1: } \epsilon = & \ \epsilon_{\theta}(x,\phi,A) + \omega \cdot (\epsilon_{\theta}(x,c,A) - \epsilon_{\theta}(x,\phi,I))\\
\text{Strategy 2: } \epsilon = & \ \epsilon_{\theta}(x,\phi,A) + \omega \cdot (\epsilon_{\theta}(x,c,A) - \epsilon_{\theta}(x,\phi,A))\\
& \ + \lambda \cdot (\epsilon_{\theta}(x,c,A) - \epsilon_{\theta}(x,c,I)) \\
\text{Strategy 3: } \epsilon = & \ \epsilon_{\theta}(x,\phi,A) + \omega \cdot (\epsilon_{\theta}(x,c,U) - \epsilon_{\theta}(x,\phi,A))\\
& \ + \lambda \cdot (\epsilon_{\theta}(x,c,A) - \epsilon_{\theta}(x,c,I)) \\
\text{Strategy 4: } \epsilon = & \ \epsilon_{\theta}(x,\phi,A) + \omega \cdot (\epsilon_{\theta}(x,c,U) - \epsilon_{\theta}(x,\phi,A))\\
\end{split}
\label{eq:CFG_Ours_supp}
\end{equation}

Table~\ref{tab:improved_video_supp} presents the comparison experimental results. From Table~\ref{tab:improved_video_supp}, Strategy 4 achieves the highest scores in Aesthetic Score and Subject Consistency. Additionally, Strategy 4 exhibits the second highest imaging quality score. Notably, Strategy 3 excels in the Motion Smoothness indicator. In conclusion, considering all four indicators, Strategy 4 demonstrates the most effective overall performance in video quality enhancement, with Strategy 3 as the next best option.

\begin{table}[ht]
\centering
\aboverulesep=0pt
\belowrulesep=0pt
\begin{small}
\begin{tabularx}{0.480\textwidth}{c|XXXX}
\toprule
Method & AS $\uparrow$ & IQ $\uparrow$ & MS $\uparrow$ & SC $\uparrow$ \\
\hline
Original  & 0.3240 & \underline{0.1965} & 0.9770 & 0.9796 \\
\hline
Strategy 1  & 0.3206 & 0.1963 & 0.9606 & 0.9536 \\
\hline
Strategy 2  & 0.3226 & 0.1964 & 0.9770 & 0.9773 \\
\hline
Strategy 3 & \underline{0.3247} & \underline{0.1965} & \textbf{0.9803} & \textbf{0.9934} \\
\hline
Strategy 4 & \textbf{0.3254} & \textbf{0.1966} & \underline{0.9797} & \underline{0.9921} \\
\hline
\bottomrule
\end{tabularx}
\end{small}
\caption{Results of video quality enhancement. AS: Aesthetic Score~\cite{Laion_5b}, IQ: Imaging quality~\cite{VBench} MS: Motion Smoothness~\cite{VBench}, SC: Subject Consistency~\cite{VBench}.}
\label{tab:improved_video_supp}
\end{table}

\subsection{Results of Video Editing}
\begin{figure*}[ht]
\centering
\includegraphics[width=0.950\textwidth]{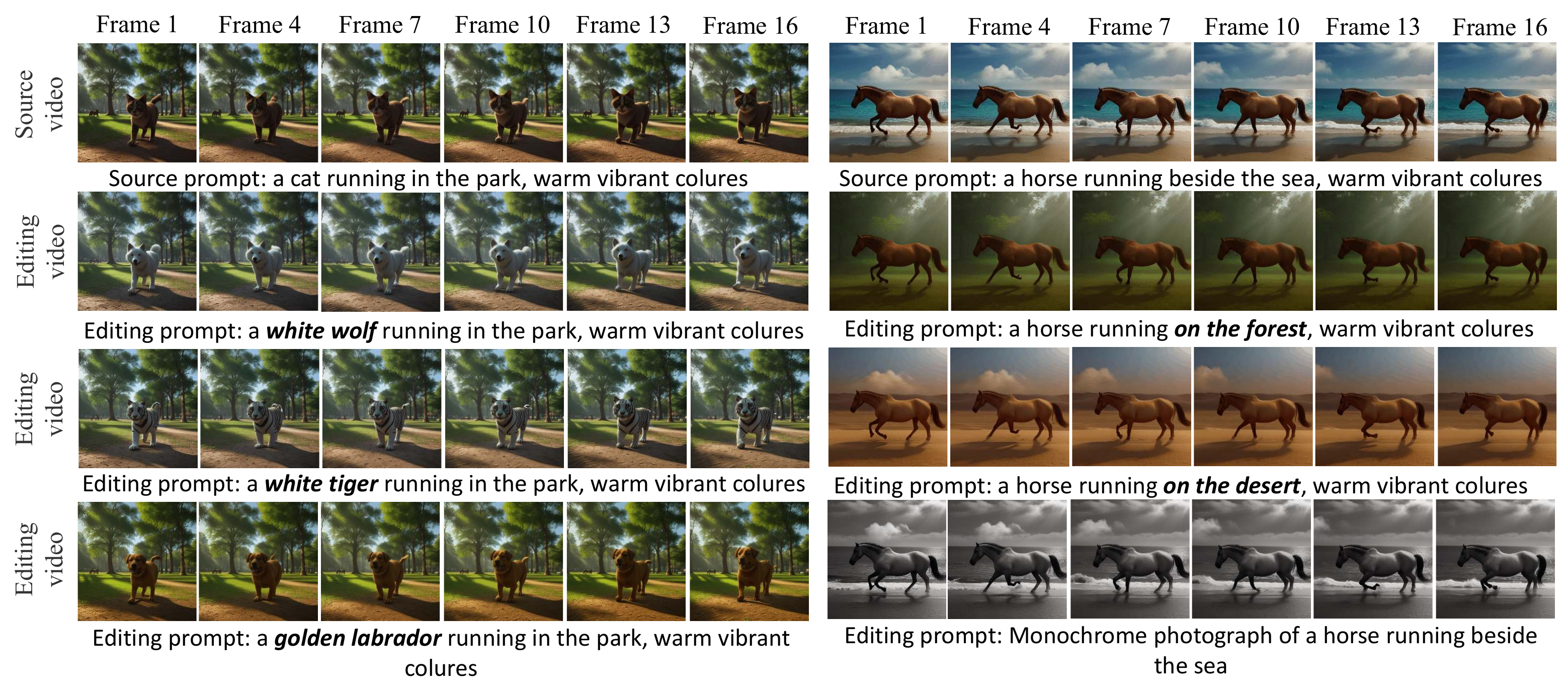}
\caption{Video editing results with AnimateDiff-Lightning.}
\label{fig:ad_ligntning_results_supp}
\end{figure*}

\begin{figure*}[ht]
\centering
\includegraphics[width=0.950\textwidth]{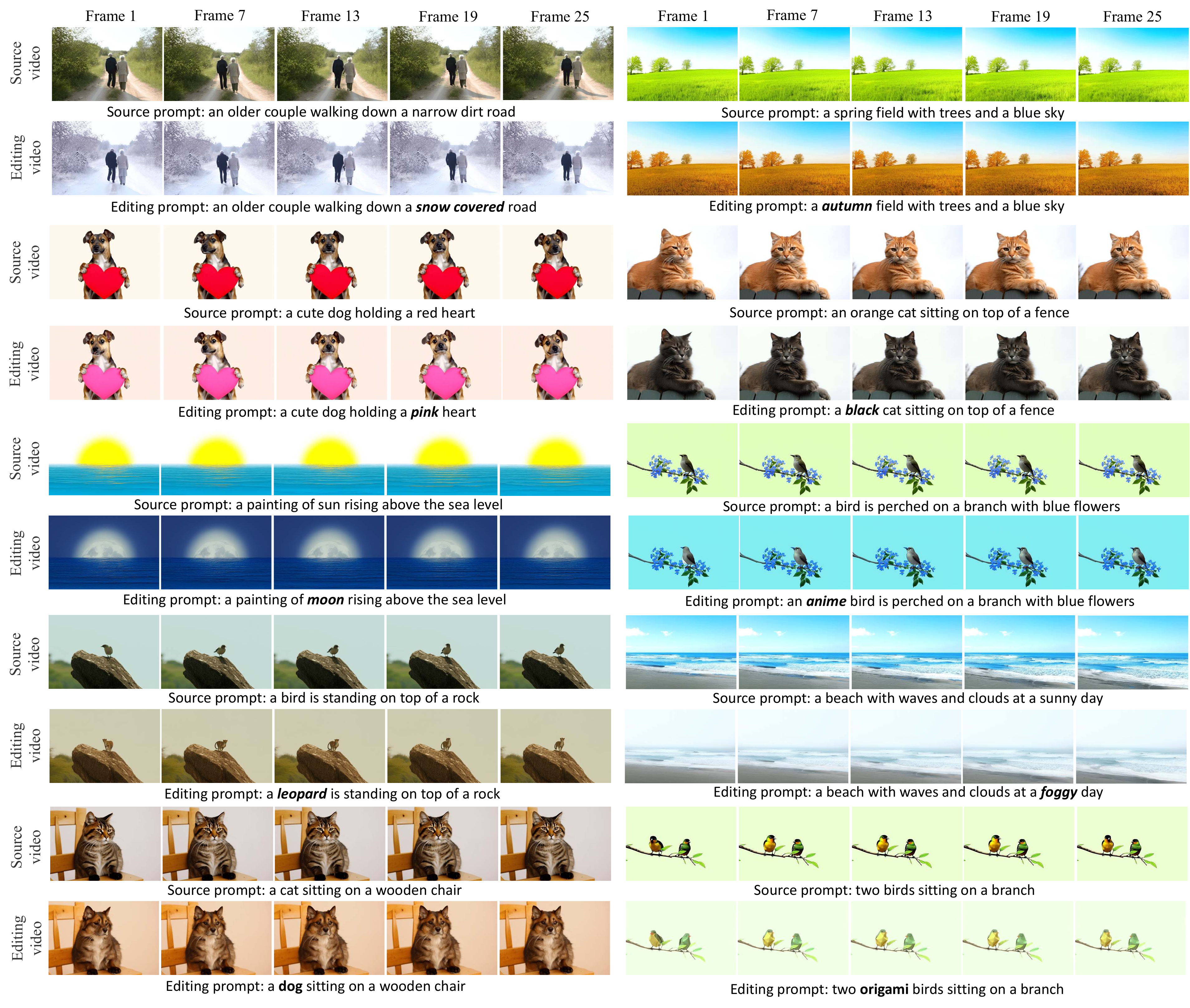}
\caption{Video editing results with CogVideoX.}
\label{fig:cogvideo_editing_results_supp}
\end{figure*}

In Figure~\ref{fig:ad_ligntning_results_supp} and Figure~\ref{fig:cogvideo_editing_results_supp}, we present additional results of applying our method to video editing on other video generation models. Table~\ref{tab:editing_result_supp} shows the results of the layers-to-edit experiment under the guidance of different percentages of information entropy. The experimental results show that a lower replacement ratio can achieve a higher CLIP score, but it will simultaneously lead to a decrease in inter-frame consistency. Conversely, full replacement reduces text consistency but improves inter-frame consistency. Taking all factors into consideration, a replacement ratio between 50\% and 65\% appears to achieve a balance between text consistency and inter-frame consistency.

\begin{table}[ht]
\aboverulesep=0pt
\belowrulesep=0pt
\begin{tabularx}{0.48\textwidth}{c|XXXX}
    \toprule
    Method & CS$\uparrow$  & CDS $\uparrow$ & MS $\uparrow$ & SC $\uparrow$ \\
    \midrule
    25\% &  \textbf{29.74} & \textbf{0.2574} & \underline{0.9787} & 0.9864\\
    50\% & 29.38 & \underline{0.2475} & \underline{0.9781} & 0.9913 \\
    65\%\textsuperscript{*} & \underline{29.54} & 0.2286 & 0.9769 & 0.9932 \\
    75\% & 29.26 & 0.2246 & \underline{0.9769} & \underline{0.9940}\\
    100\% & 27.94 & 0.1888 & 0.9766 & \textbf{0.9946} \\
    \bottomrule
\end{tabularx}
\caption{Quantitative results of video editing on AnimateDiff. CS: Clip Score~\cite{clip}, CDS: Clip Directional Similarity~\cite{clip,StyleGAN-NADA}, MS: Motion Smoothness~\cite{VBench}, SC: Subject Consistency~\cite{VBench}.}
\label{tab:editing_result_supp}
\end{table}

\subsection{Combined with masking method}
\begin{figure}[ht]
\centering
\includegraphics[width=0.49\textwidth]{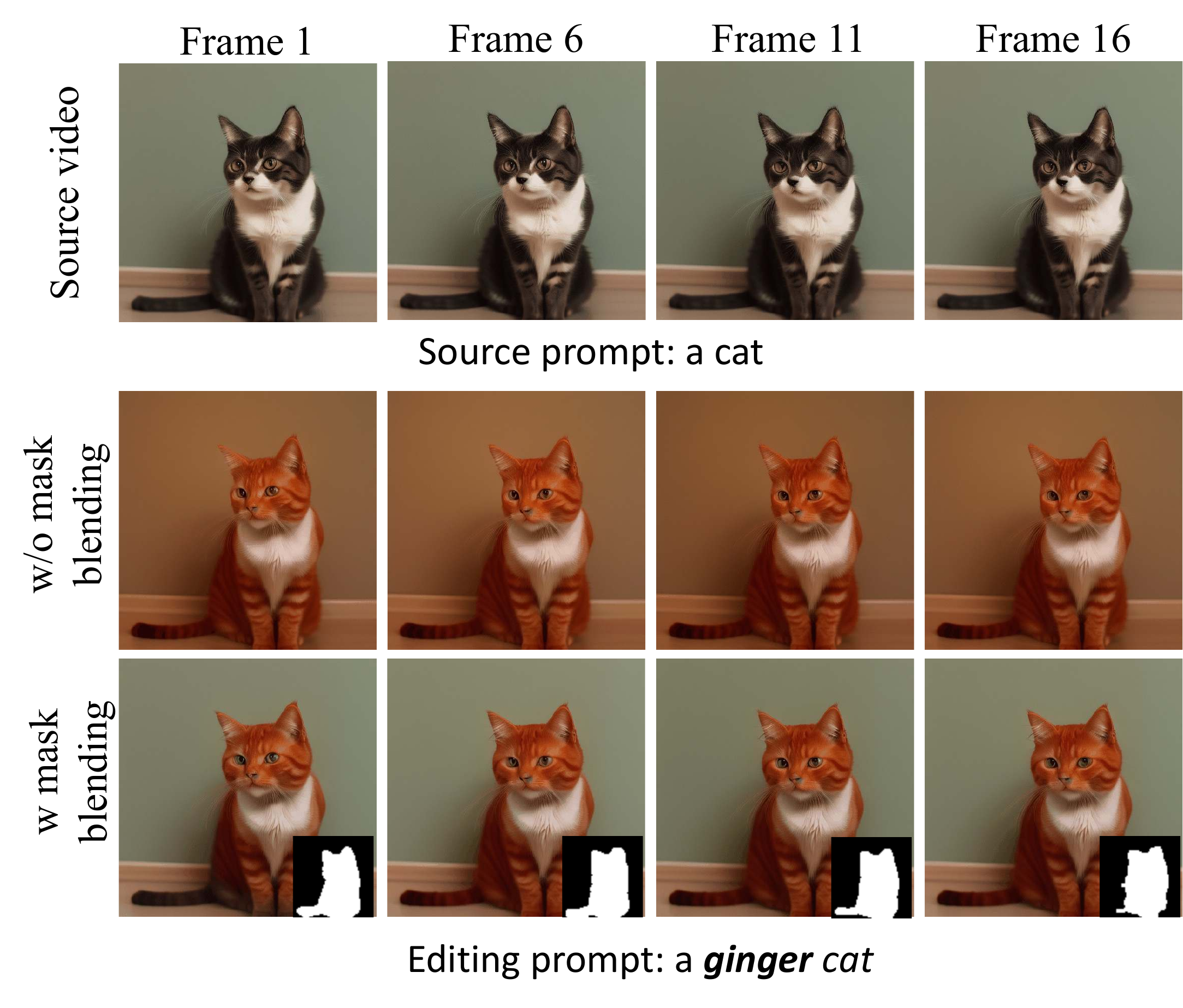}
\vspace{-1em}
\caption{Video editing results under mask guides. }
\label{fig:mask_guiden_suppl}
\vspace{-1em}
\end{figure}

Model bias and attribute leakage in diffusion models can sometimes result in background changes in edited videos. Our method can be combined with masking techniques to achieve background preservation, as shown in Figure~\ref{fig:mask_guiden_suppl}.

\end{document}